\documentclass[journal,twoside, web]{ieeecolor}
\usepackage{jsen}
\usepackage{cite}
\usepackage{amsmath,amssymb,amsfonts}
\usepackage{algorithmic}
\usepackage{graphicx}
\usepackage[caption=false,font=footnotesize]{subfig}
\usepackage{textcomp}
\usepackage{wrapfig}

\usepackage{xurl}      

\usepackage{xcolor}

\def\BibTeX{{\rm B\kern-.05em{\sc i\kern-.025em b}\kern-.08em
    T\kern-.1667em\lower.7ex\hbox{E}\kern-.125emX}}
\definecolor{abstractbg}{rgb}{0.89804,0.94510,0.83137}
\makeatletter
\let\ps@titlepagestyle\ps@headings
\makeatother

\begin{document}
\title{Direct-Ink-Writing of Soft Thin Film Force-Sensitive Resistors with Velostat and Printed-Carbon Sensing Elements}

\author{Mostafa Mo. Massoud, \IEEEmembership{Graduate Student Member, IEEE}, Paulo H.T.F Alves, \IEEEmembership{Graduate Student Member, IEEE}, Timothy Kennedy,  \IEEEmembership{Student Member, IEEE}, and Jacqueline Libby, \IEEEmembership{Member, IEEE}
\thanks{This work was supported by the US National Science Foundation under Grant 2502197. (\emph{Corresponding author: Jacqueline Libby}) }
\thanks{Mostafa Mo. Massoud, Paulo H.T.F Alves, and Jacqueline Libby are with the Department of Mechanical Engineering, Stevens Institute of Technology, Hoboken, NJ 07030 USA.
    (e-mail: mmassoud@stevens.edu;pteixeir@stevens.edu; tkennedy1@stevens.edu; jlibby@stevens.edu) }}

\IEEEtitleabstractindextext{%
\fcolorbox{abstractbg}{abstractbg}{%
\begin{minipage}{\textwidth}%
\begin{wrapfigure}[15]{r}{3.1in}%
\includegraphics[width=3in]{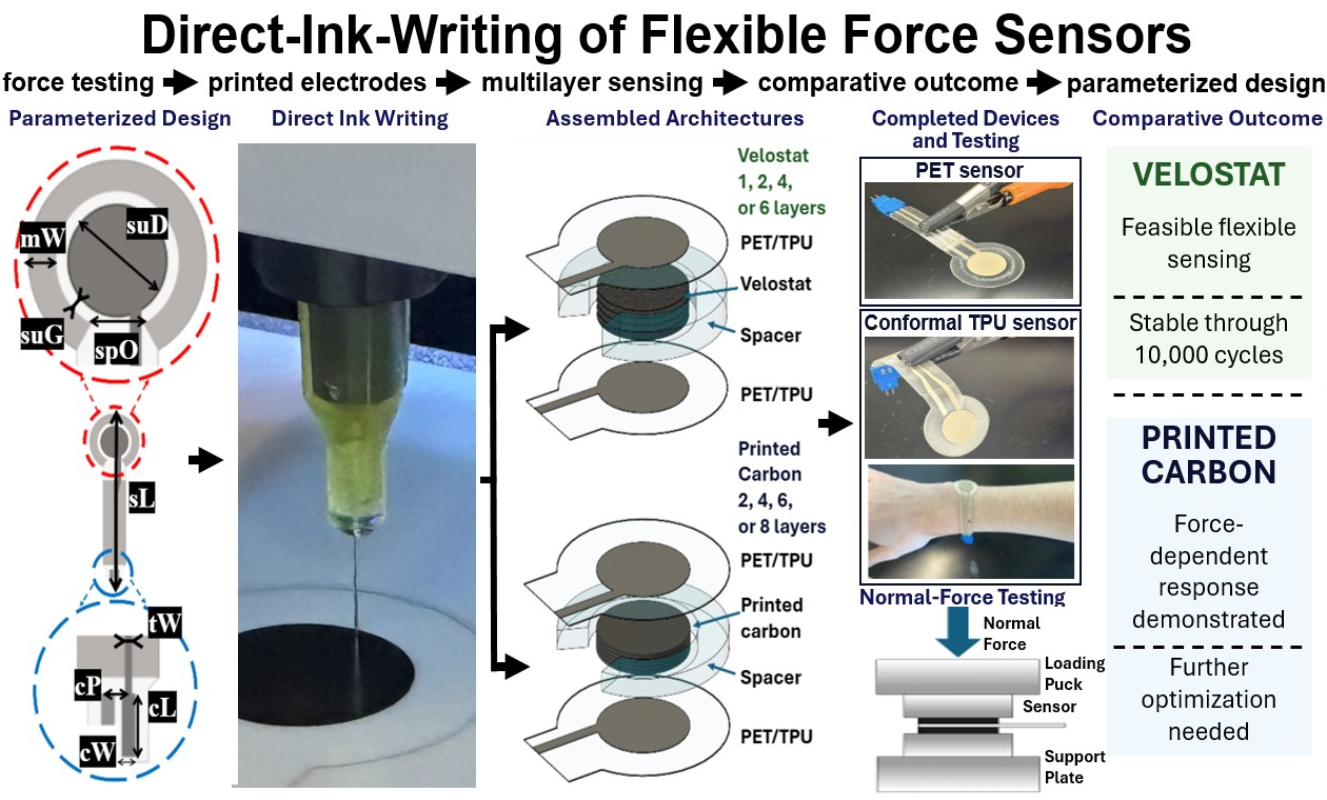}%
\end{wrapfigure}%

\begin{abstract}
Flexible resistive force sensors are useful for wearable, rehabilitation, and human--machine interfaces, but their performance depends on sensing-material architecture and fabrication. This work presents a parameterized Voltera NOVA-based direct-ink-writing workflow using printed silver electrodes with multilayer Velostat sensing elements on PET and TPU substrates, together with an exploratory fully printed carbon architecture. Velostat sensors containing 1, 2, 4, and 6 sensing layers exhibited strongly nonlinear force--voltage responses, with the greatest sensitivity at low forces. Best-fit straight-line $R^2$ values ranged from 0.475 to 0.641, area-based hysteresis from 3.90\% to 7.47\%, and 5-min voltage drift from 1.9\% to 7.3\%, with no consistent monotonic relationship between layer count and these metrics. Selected PET6 and TPU6 devices showed less than 0.05~V change in peak output between the first and final 100 cycles of a 10,000-cycle test. Printed-carbon devices demonstrated the feasibility of a more fully printed architecture, but configurations with six or fewer layers saturated below approximately 0.5~N or were nonfunctional, while the eight-layer device responded primarily below approximately 2~N. These results establish the feasibility of the printed-electrode Velostat workflow while identifying printed sensing materials as a key area for further development.
\end{abstract}

\end{minipage}}}

\maketitle
\section{Introduction}\label{sec.Introduction}

\IEEEPARstart{F}{lexible} force sensors are important in wearable systems, rehabilitation devices, soft robotics, and human–machine interfaces because they enable contact loads to be measured at curved, compliant, or moving interfaces~\cite{zhu2022recent, luo2023technology}. Recent developments in flexible bioelectronics and wearable
sensing further emphasize the importance of mechanically compliant materials and flexible electronic platforms for stable,
human-compatible sensing interfaces \cite{Wu2025-CBS,gong2026JIOHS}. In this paper, force denotes the total normal load transmitted through the sensing region, whereas pressure denotes force normalized by a defined contact area. Because the devices developed here are calibrated directly against an applied force rather than an area-normalized or spatially distributed pressure, they are referred to throughout as force sensors.

Flexible force sensors can employ piezoresistive, piezoelectric, capacitive, optical, triboelectric, or magnetic transduction~\cite{zhu2022recent, luo2023technology}. This study focuses on piezoresistive sensing because changes in electrical resistance can be measured under both static and cyclic loading using a comparatively simple voltage-divider circuit. In a piezoresistive sensor, mechanical deformation, changes in conductive pathways, or changes in interfacial contact alter the resistance of the sensing element, which is converted to an output voltage by the readout circuit. Piezoresistive architectures can be inexpensive and mechanically compliant, but their performance may be affected by nonlinear response, hysteresis, time-dependent drift, and variability associated with the sensing material, electrode contact, and assembly conditions~\cite{zhao2025flexible,matute2017characterization}.

Force-sensor performance must be evaluated under defined mechanical, electrical, environmental, and temporal conditions. The principal quantities considered in this study are sensitivity, calibration nonlinearity, hysteresis, drift, response time, and cyclic stability. Sensitivity is the change in a specified sensor indication relative to the corresponding change in force, whereas hysteresis is the difference between loading and unloading outputs at the same applied force. Drift describes a time-dependent change in output under a constant load. Cyclic stability describes changes in performance following prolonged repeated loading made under the same conditions. The settling criterion used to calculate response time must also be stated explicitly~\cite{nie2024mechanically, cao2021six,matute2017characterization}.

Flexible force sensors can be fabricated using microfabrication, screen printing, inkjet printing, coating and lamination, vision-based systems, molding, and digitally controlled direct-write processes~\cite{olowo2023design,maddipatla2020recent, han2023review, goh20213d, Fan2025-CyborgBionicSyst}. Microfabrication can provide high feature resolution and mature batch processing but generally requires substantial equipment and process infrastructure. Screen printing is well suited to repeated, high-throughput deposition but normally requires a dedicated screen for each pattern. Inkjet printing is maskless and material-efficient but imposes strict requirements on ink viscosity, particle size, surface tension, wetting, and nozzle operation.
Direct ink writing is an established extrusion-based process in which a digitally defined toolpath is used to deposit functional inks without a patterned mask~\cite{Lewis2006AFM, Saadi2022AM}. DIW can accommodate comparatively viscous inks and facilitate design changes and multilayer deposition, although its performance depends on ink rheology, nozzle dimensions, wetting, layer registration, curing, and interlayer adhesion~\cite{goh20213d, tay2023direct, rocha2020direct, Saadi2022AM}.
Vision-based systems provide spatial maps of contact shape and indentation depth, but their camera, lighting, and calibrated image processing imply a bulkier system that is harder to integrate into thin, conformable devices~\cite{Fan2025-CyborgBionicSyst}.
In the present work, DIW is used to print conductive electrodes and, for the printed-carbon architecture, the sensing layers.

Direct printing and DIW have previously been applied to tactile and force- or pressure-sensitive devices. Vatani et al.~\cite{Vatani2015AdditiveManufacturing} fabricated compliant multilayer piezoresistive tactile sensors through a hybrid DIW/stereolithography process. Van Dommelen et al.~\cite{Dommelen2020IEEESensors} reported fully 3D-printed mechanical pressure sensors based on multiple sensing mechanisms, and Baghini et al.~\cite{Baghini2020FLEPS} directly wrote graphite-based soft piezoresistive pressure sensors. These studies preclude a claim that the present device is the first force sensor fabricated using any DIW process. The contribution of the present study should instead be framed around its Voltera NOVA-based fabrication workflow (rather than a more expensive bioprinting process), parameterized geometry, selected commercial materials, and comparative characterization of the resulting force sensors.

Velostat is a carbon-loaded polyethylene sheet whose electrical resistance changes under compression and has been widely used in low-cost resistive force sensors~\cite{Prete2001RSI, Cesteros2023IEEESensorJournal, kalantari2011ASME, vehec2020ISSE}. Its device-level behavior is influenced by electrode contact, preload, geometry, material relaxation, and loading history, which can result in nonlinear response, hysteresis, and drift. Bartunek et al.~\cite{bartunek2025Sensors} evaluated sensors containing 1, 3, 4, 5, 10, 20, and 30 Velostat layers and found that multilayer designs reduced force-estimation error relative to a one-layer design and a commercial force-sensing resistor under static, cyclic, and finger-loading evaluations. However, that study focused primarily on measurement error and did not comprehensively examine hysteresis, time-dependent output, response behavior, high-cycle stability, or PET-versus-TPU substrate effects within a common printed-electrode architecture. These remaining questions motivate the multilayer Velostat investigation in the present study.

Printed carbon composites can exhibit piezoresistive behavior, but their performance depends on conductive-filler concentration, matrix properties, microstructure, rheology, curing, and printed geometry. Recent DIW work using MWCNT/PDMS nanocomposite inks illustrates that printability and electromechanical response must be optimized together when developing a pressure- or force-sensitive material~\cite{Sidhu2025NanoscaleAdvances}. In contrast, ACI SC1502, the carbon material used in this study, is marketed as a stretchable carbon conductor for printed circuitry rather than as a purpose-designed force-sensing ink~\cite{sc1502_2023}. The printed-carbon device is therefore evaluated here as an exploratory test of whether stacked conductor prints can produce a reproducible force-dependent resistance response within the proposed architecture.

Based on the preceding literature, the unresolved question is not whether
DIW can be used to fabricate force-sensitive devices, but how the proposed
Voltera NOVA-based workflow performs with the selected sensing materials,
layer arrangements, and substrates. This study addresses three research
questions: (1) How does increasing the number of Velostat layers from one
to six affect the force-dependent response and selected performance measures
of sensors with printed electrodes? (2) How does replacing a PET substrate
with TPU affect the response of otherwise comparable multilayer Velostat
sensors? (3) Can a more fully printed force-sensor architecture, in which
both the conductive electrodes and sensing element are fabricated using the
Voltera NOVA, produce a measurable force-dependent response using a
commercial printed-carbon conductor?

To address these questions, this paper: (1) describes a parameterized
CAD-to-DIW and assembly workflow for flexible resistive force sensors;
(2) characterizes Velostat sensors containing 1, 2, 4, and 6 sensing layers
on PET and TPU substrates; (3) evaluates an exploratory printed-carbon
architecture as a demonstration of a more fully printed sensor fabricated
using the Voltera NOVA; and (4) quantitatively evaluates force-dependent
sensitivity, calibration nonlinearity, hysteresis, drift, and cyclic
stability using a voltage-divider readout. The limitations of the present experimental system for resolving sensor response time are also discussed. Section II describes the materials, design, fabrication, assembly, and electrical readout; Section III presents the characterization procedures and results; Section IV discusses the experimental findings, comparisons, and limitations; and Section V summarizes the conclusions and future work.


\section{Materials and Methods}\label{sec.Materials_and_Methods}

\subsection{Materials}

\begin{figure}[!t]
\centering
\includegraphics[width=1\linewidth]{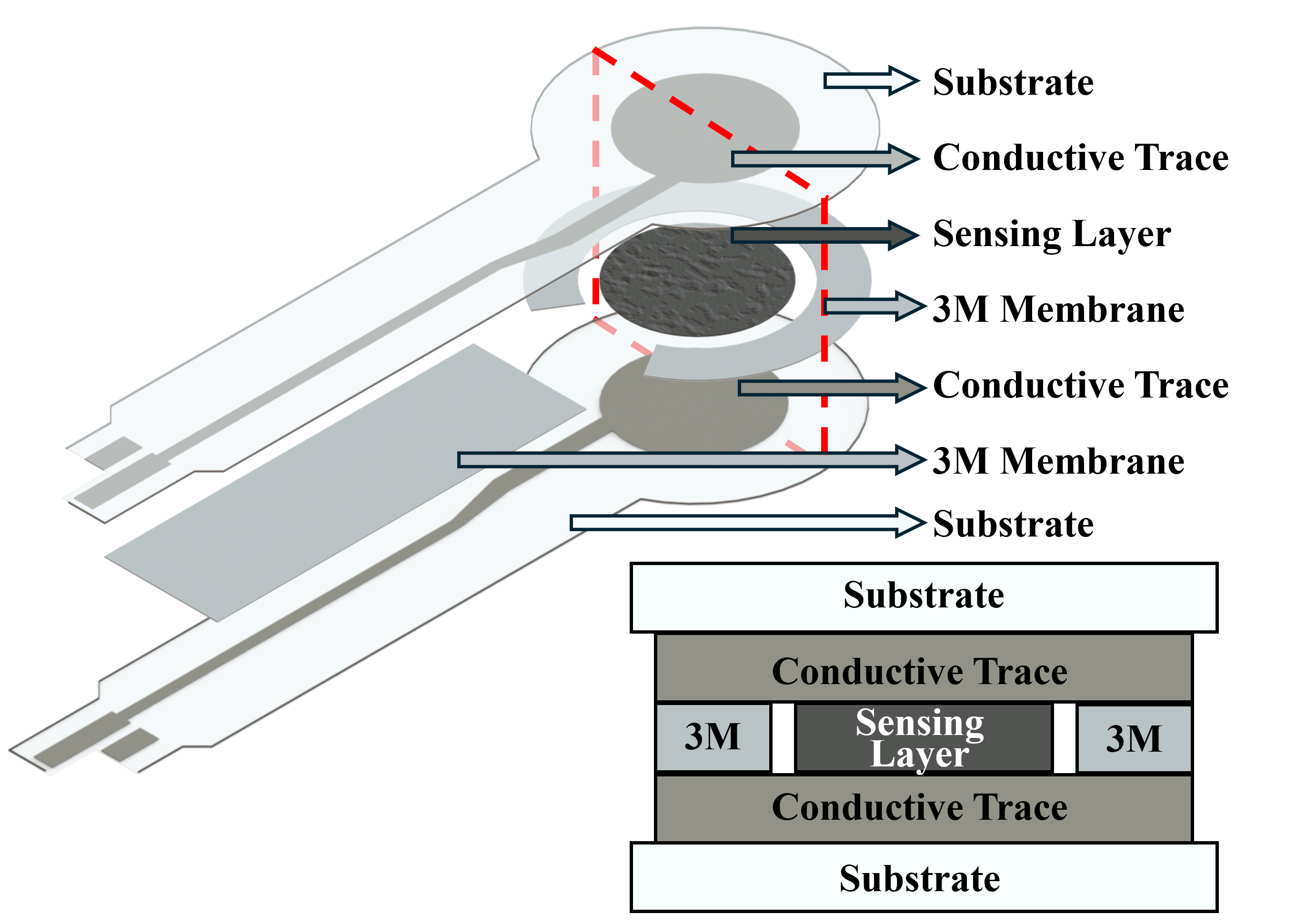}
\caption{Sensor exploded view with assembly block diagram.}
\label{fig.explodedview}
\end{figure}

A family of through-mode force-sensitive resistor sensors is presented, with each version varying in materials and manufacturing processes.  The different versions within the family share a similar layout, depicted in Fig.~\ref{fig.explodedview}.  
The substrates form the outer layer of the sensor and serve to insulate, protect, and support the sensing elements. Accordingly, they are typically made from materials with high insulating capacity to shield the sensor from external influences, strong chemical and mechanical resistance to ensure durability, and physical characteristics that align with the sensor’s purpose, such as rigidity, flexibility, softness, or stretchability.
Conductive layers act as the interface between the sensing unit and the readout system. These layers must exhibit high electrical conductivity to guarantee that the sensor’s output is not distorted or attenuated by resistive losses during signal transmission.
The sensing layer changes resistance when a compressive force is applied. As the sensing layer is compressed, conductive particles within the material come into greater contact, thereby lowering the resistance. The material utilized drives the sensor properties with regard to sensitivity, hysteresis, drift, and other metrics.
The assembly layer is responsible for attaching components as well as creating the sensor vent, which is a space that allows airflow inside the sensing unit, thus allowing proper function. If airflow is not properly addressed, trapped air inside the sensor can behave like a spring, preventing full compression and introducing unwanted mechanical stiffness. If air is expelled without a way to re-enter, a partial vacuum may form, creating unintended preload and altering the sensor’s baseline response.

Each material in the sensor assembly influences the sensor's final properties, and thus must be carefully selected.
For this research, BOPET (Biaxially Oriented Polyethylene Terephthalate - Normandy Coating Arcophane AA STS)  and TPU (Thermoplastic Polyurethane - Intexar TE-11C) were selected as substrates, ACI SS1109 stretchable silver ink was selected for printing conductive traces, a pressure-sensitive conductive sheet (Velostat) and the ink ACI SC1502 stretchable carbon ink were selected as materials for the sensing layers, and the membrane (3M 7945MP) was selected for the assembly layer.

The Intexar TE-11C TPU is a stretchable bilayer film with high recovery and melt adhesive layers optimized for high stretch smart applications, designed for compatibility with stretchable electronic ink printing \cite{intexar_te11c_2025}.
The Arcophane AA STS BOPET is a transparent two-sided coated film with an acrylic treatment that improves bonding property for material deposition and avoids shrinkage during the curing process \cite{arcophane_aa_sts_pet_2025}.
The stretchable printed silver conductor (SS1109) is a silver filled conductor that demonstrates excellent volume resistivity ($<4.5 * 10^{-5} \Omega * cm$) and excellent elongation ($>200\% $) \cite{ss1109_2025}.
The stretchable printed carbon conductor (SC1502) has a higher volume resistivity when compared to the SS1109 ink ($<0.5 \Omega * cm$) and also exhibits excellent elongation ($>200\% $) \cite{sc1502_2023}.
The pressure-sensitive conductive sheet (Velostat) has good flexibility and a volume resistivity of ($<500 \Omega * cm$) and a surface resistivity of ($<31,000 \Omega/sq$) \cite{pressure_sensitive_conductive_sheet_2025}.
The 3M 7945MP membrane is a clear bonding tape that uses an acrylic adhesive capable of withstanding temperatures from -40°F to +400°F and also has good flexibility \cite{membrane_switch_spacer_7945mp_2025}.

\subsection{Design}
\normalcolor
This research proposes two types of additive-manufactured sensors: Velostat and Printed. In the Velostat sensor, the conductive traces (ACI SS1109 Stretchable Silver Ink) were printed on both PET and TPU substrates, and for the sensing layer, a pressure-sensitive conductive sheet (Velostat) was utilized. In the printed sensor, both the conductive (ACI SS1109 Stretchable Silver Ink) and sensing traces (ACI SC1502 Stretchable Carbon Ink) were printed on PET substrates.

\begin{figure}[!t]
\centering
\includegraphics[width=1\linewidth]{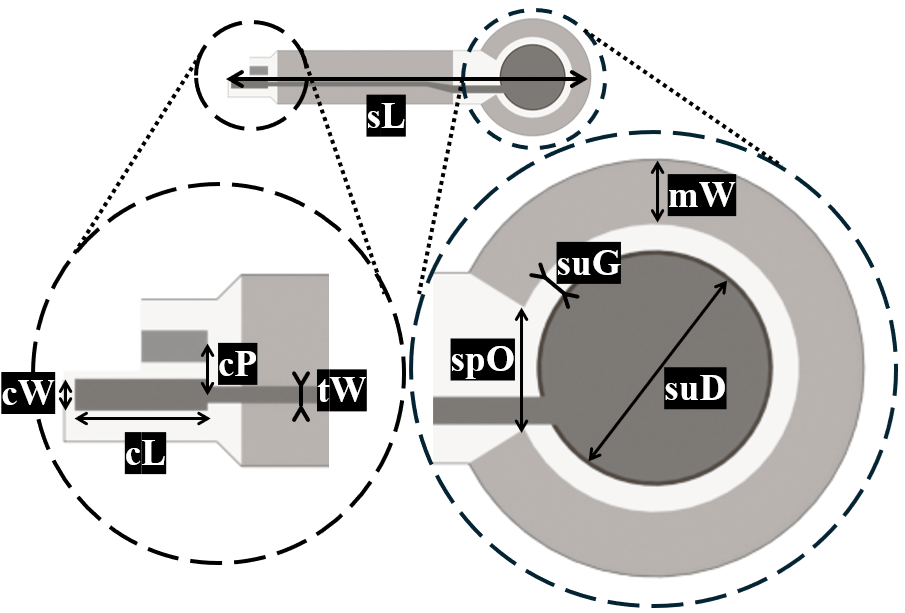}
\caption{Sensor dimensions}
\label{fig.sensorDesign}
\end{figure}

Each sensor layer was designed in SolidWorks rather than PCB software to enable parametric design for future customization and future integration with 3D structures onto which these sensors will eventually be mounted. The sizing dimensions are parameterized as shown in Fig.~\ref{fig.sensorDesign}. The parameters include the sensor length (sL), the sensing unit diameter (suD), the sensing unit gap (suG), the membrane width (mW), the spacer opening (spO), the trace width (tW), the connector length (cL), the connector width (cW), and the connector pitch (cP). The parameter values are given in Table~\ref{tab.sensorvariables}. cW, cP, and cL were chosen based on the sizes of a 2 Position Flat Flexible Cable (FFC) connector Header with a 0.1" pitch to allow good connectivity and avoid issues caused by misalignment. tW was selected according to the optimal trace width for the selected nozzle and trace space in order to avoid trace gaps and/or overlapping traces. mW was selected based on the available sizes for a hole punch tool, suD was selected to be the same size as a typical human index finger width, and suG was selected as 0 for Velostat sensors to facilitate assembly, while 0.5 for printed sensors to avoid membrane misalignment issues. sL and sPO were arbitrarily chosen.

\begin{table}[]
\begin{center}
\caption{Sensor variables}
\label{tab.sensorvariables}
\begin{tabular}{|c|c|c|}
\hline
\textbf{Name}                  & \textbf{Abbreviation} & \textbf{Dimensions(mm)} \\ \hline
Sensor length         & sL          & 68.5           \\ \hline
Membrane Width        & mW          & 3.5      \\ \hline
Sensing Unit gap      & SuG         & 0.5(P) 0(V)      \\ \hline
Sensing Unit Diameter & SuD         & 12       \\ \hline
Spacer Opening        & spO         & 6.4      \\ \hline
Trace Width           & tW          & 0.9      \\ \hline
Connector Pitch       & cP          & 2.5      \\ \hline
Connector length      & cL          & 7        \\ \hline
Connector Width       & cW          & 1.7      \\ \hline
\end{tabular}
\end{center}
\end{table}

\begin{figure}[!t]
\centering
\includegraphics[width=1\linewidth]{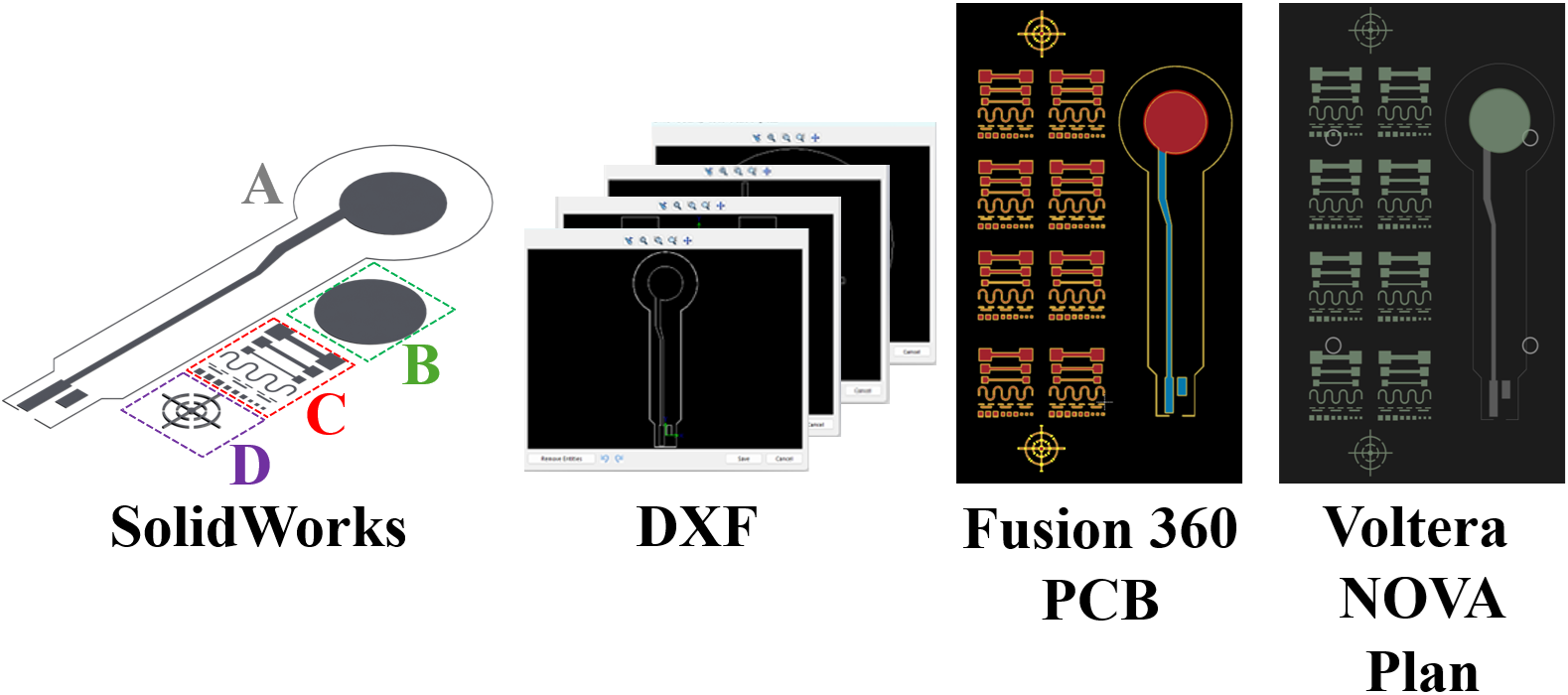}
\caption{Pipeline to design sensors and input them into the Voltera NOVA, where A represents the conductive trace, B represents the sensing trace, C represents the calibration pattern, and D represents the alignment fiducial. The multilayer sensors are designed and assembled in SolidWorks, and then exported to DXF surface files. Traces are filled in Fusion 360 PCB, and then exported to Gerber files, which are used as inputs to the Voltera NOVA software.}
\label{fig.pipeline}
\end{figure}

After choosing parameter values, the conductive, sensing, cutting, and alignment traces were designed as sketches and extruded by 0.1mm to generate a 3D structure inside SolidWorks. Conductive traces will delimitate where the conductive material is being deposited, the sensing traces will delimitate where the sensing material is being deposited, the alignment traces will be used to align printed traces for multilayer prints, and the cutting trace will serve as a guide for cutting the substrates during assembly post-printing.
Once each drawing is extruded into a 3D structure, a surface of that structure is selected and exported as a DXF file.  The DXF file contains the outer perimeter of a given surface. This file is imported into Fusion 360 PCB workspace, maintaining the perimeters as "board outline" layer types, which allows all shape details to be maintained in their original form.  Then the "Polygon Pour From Outline" tool is used to fill in the perimeters for all shapes. After all traces are filled, the file is saved as a Gerber file. The Gerber file, which is the industry standard for describing printed circuit board layers, is the input to the Voltera NOVA software. The process from SolidWorks parts to Voltera NOVA software is illustrated in Fig.~\ref{fig.pipeline}, where A represents the conductive trace, B represents the sensing trace, C represents the calibration pattern, and D represents the alignment fiducial.

The design also considered a variation of quantities in sensing layers (x) and assembly layers (y). If the sensing layer height is higher than the assembly layer height, the sensor will be pre-loaded, which allows current to pass through even in unloaded stages.
Therefore, the equation  $y = \left\lceil 2x \right\rceil$ was used to calculate the number of assembly layers (y) for printed and Velostat sensors to ensure that sensors do not have a substantial amount of preloading.

\subsubsection{Velostat Sensor Design}
For the Velostat sensor, an assembly of the conductive trace together with the cutting trace was created, and the assembly's top surface was exported into a 2D DXF file. Inside Fusion 360 PCB workspace, the traces in the DXF file were filled, and calibration patterns were added for printing calibration purposes.

\subsubsection{Printed Sensor Design}
For the printed sensor, two different DXF files had to be generated: one for the conductive layer and one for the sensing layer.  The conductive layer has conductive, cutting, and alignment traces, and the sensing layer has sensing and alignment traces. To allow for alignment of the conductive and sensing layers, both files were assembled together in SolidWorks, using the alignment traces across both layers as fiducials. Then one of the layers was suppressed, allowing the creation of the DXF file without interference, and the same process was repeated for the other layer. Inside Fusion 360 PCB workspace, the DXF files were filled, and patterns were added for calibration purposes.



\subsection{Direct Ink Writing}



\begin{figure}[!t]
\centering
\includegraphics[width=1.00\linewidth]{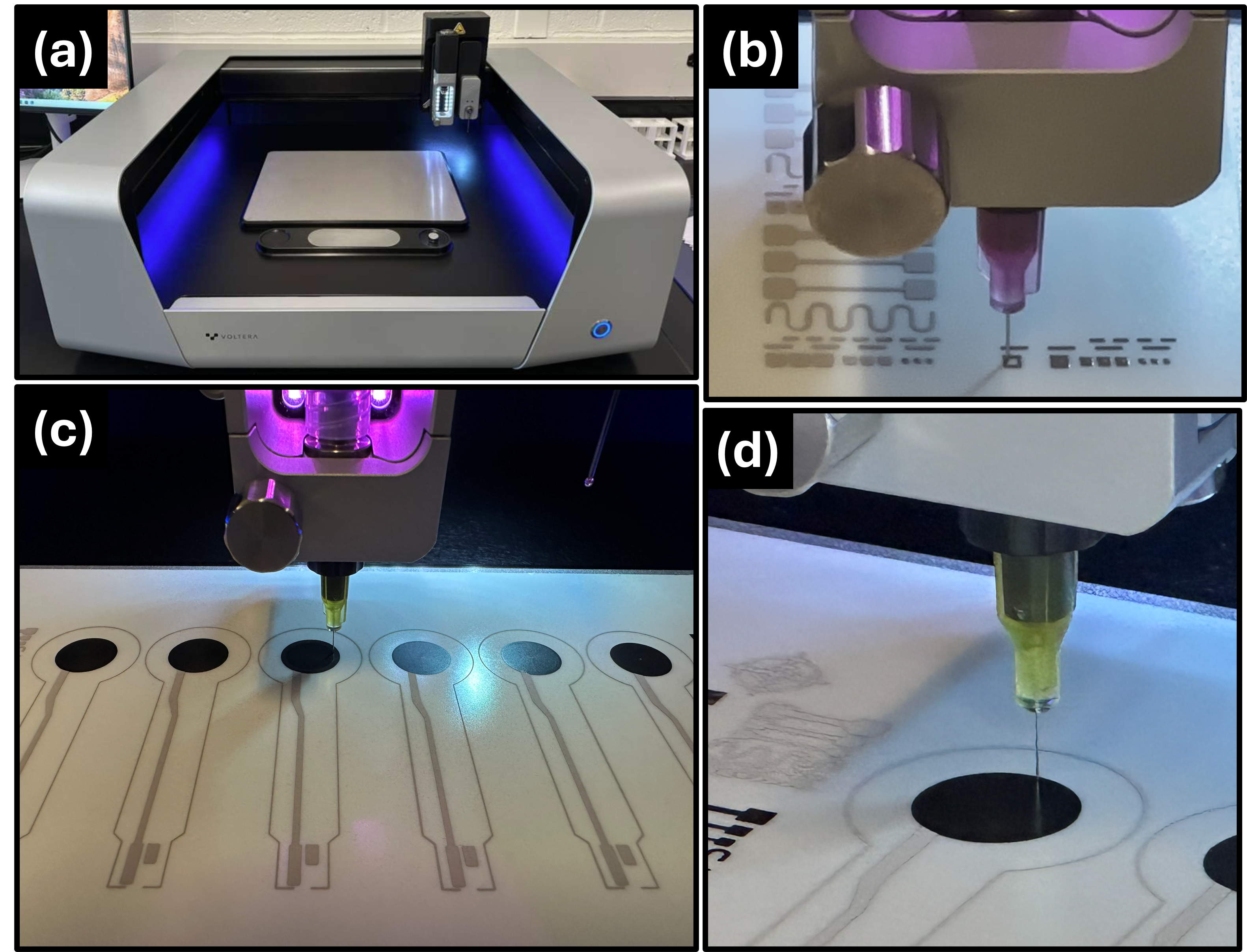}
\caption{Sensor printing process. (a) Voltera NOVA. (b) Fine-tuning calibration of ink settings with on-substrate patterns. (c) A large group of printed sensors, demonstrating the scalability of the process. (d) Close-up of a nozzle depositing a third layer of ink onto the TPU substrate.}
\label{fig.volteraNova}
\end{figure}

The additive manufacturing process, performed utilizing the Voltera NOVA (Fig. \ref{fig.volteraNova}a), starts with loading the Gerber file into the Voltera NOVA software. The substrate and ink are then selected, and a tool path is generated based on the specific settings for each ink, substrate, and nozzle combination. After aligning the substrate and assembling the ink and nozzle into the dispenser, the ink settings were calibrated using the calibration plate as a testing surface to ensure a proper flow. A second round of fine-tuning calibration was performed directly on the substrate (Fig.~\ref{fig.volteraNova}b) by printing patterns of basic shapes to ensure the shape and flow were optimal before starting to print the sensors. These calibration patterns are pre-loaded with the Gerber file, as shown in Fig.~\ref{fig.pipeline}. After successfully printing three complete calibration patterns, the sensors were printed (Figs.~\ref{fig.volteraNova}c and d).

\subsection{Velostat Sensor Manufacturing}\label{subsec.velostat}
\begin{figure}[!t]
\centering
\includegraphics[width=1\linewidth]{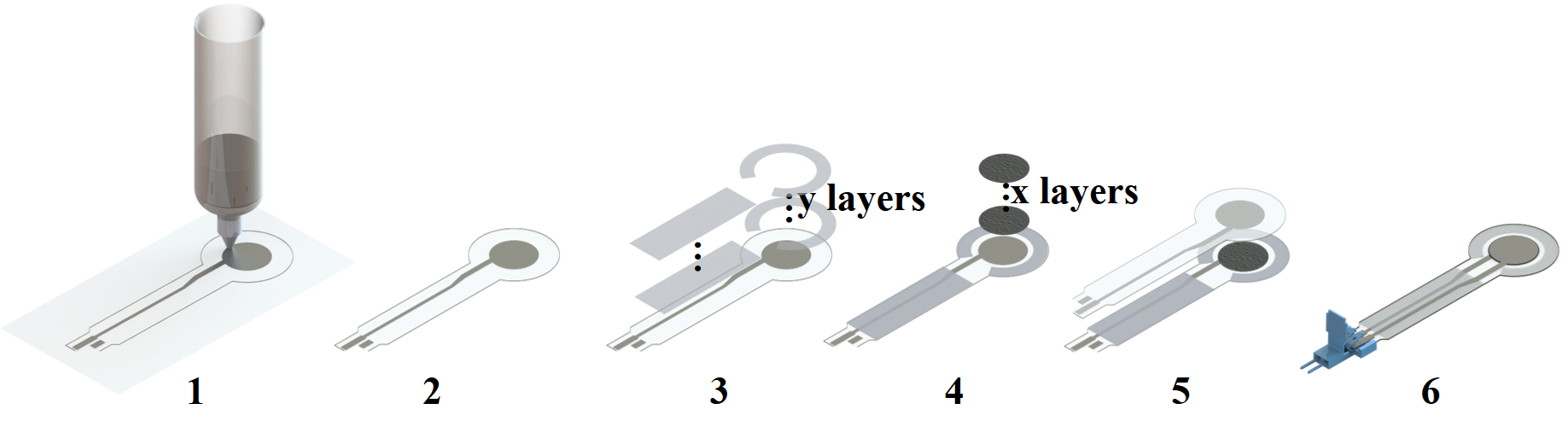}
\caption{Velostat sensor detailed manufacturing process}
\label{fig.velostatManufacturing}
\end{figure}

The Velostat sensor manufacturing process (Fig.~\ref{fig.velostatManufacturing}) started by loading the Gerber file for conductive, cutting, alignment, and calibration traces into the Voltera NOVA. Then a calibration process was performed for the conductive ink SS1109, the nozzle Nordson EFD 30GA (150 $\mu m$ diameter), and the substrates TPU or PET. Small-size nozzles (150 - 200 $\mu m$ diameter) allow printing thin layers with high precision. Variables determined by the calibration process can be found in Table.~\ref{tabInkSettings}. After calibration, the conductive traces were printed (Fig.~\ref{fig.velostatManufacturing} Step 1) and then the material was cured at 135°C for 5 minutes.

\begin{figure}[!t]
\centering
\includegraphics[width=.75\linewidth]{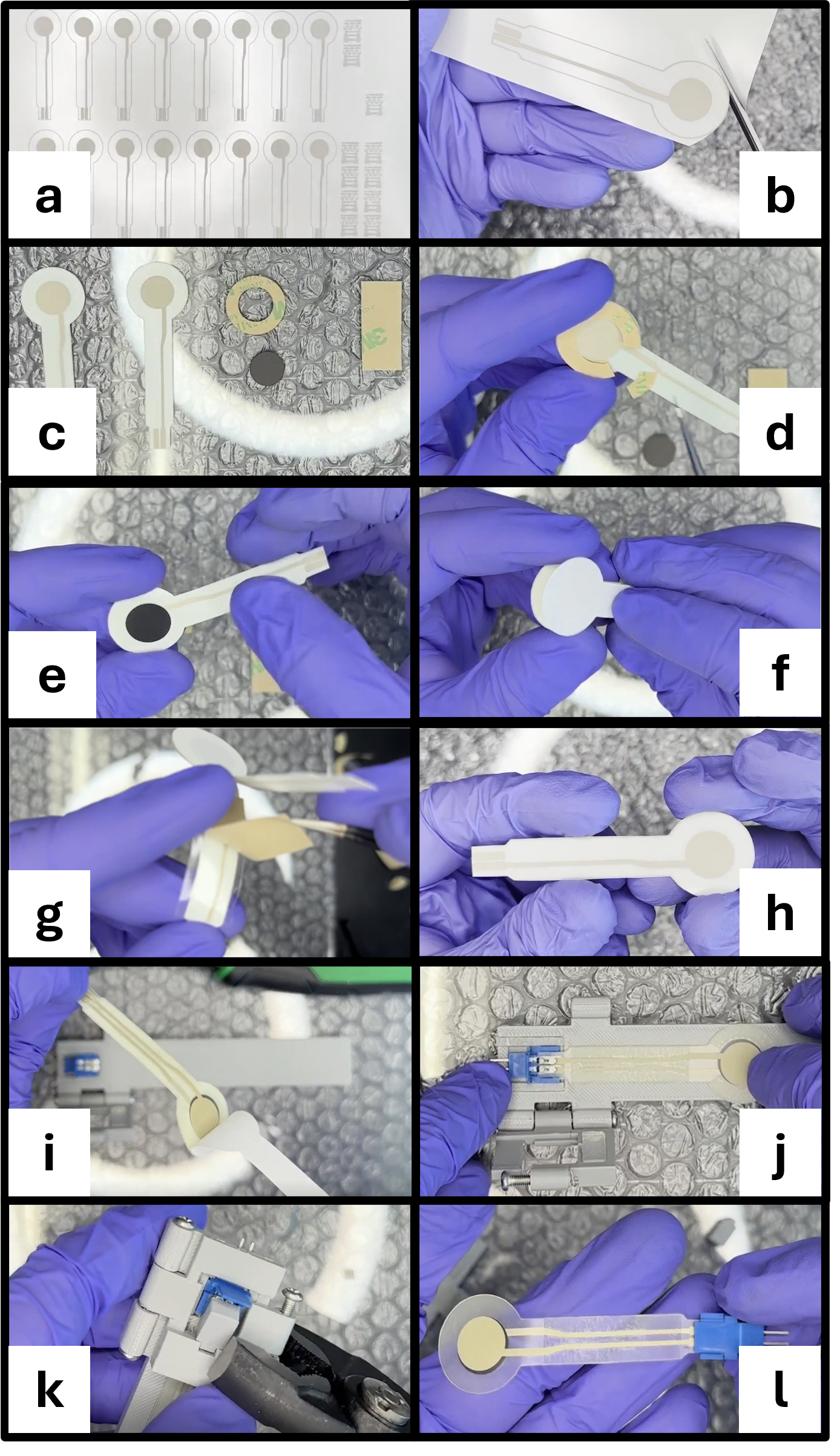}
\caption{Assembly process for the Velostat sensor. (For the printed sensor, step (e) is removed.) (a) A set of conductive traces on a sheet of TPU substrate. (For the printed sensor, the resistive carbon traces would also be printed.) (b) Cutting along cutting traces. (c) All parts of a sensor already cut (TPU, 3M membrane, Velostat). (d) Attaching a circular 3M membrane to the top substrate with print. (e) Placing Velostat layers inside the membrane. (f) Attaching the second substrate with print. (g) Attaching a rectangular membrane to the sensor legs with contact leads. (h) The assembled sensor. (i) Peeling the TPU backing. (j),(k) Installing the FFC connector with the aligning tool. (l) Final sensor.}
\label{fig.sensorAssembly}
\end{figure}

After curing, the sensor was assembled. A scissor was used to cut along the cutting traces, giving the sensor its final shape (Step 2). In step 3, the y layers of spacer material were cut and assembled, followed by Step 4, which is the cutting, alignment, and addition of x layers of Velostat. In Step 5, another layer of substrate with printed conductive ink was added to the top (with conductive traces facing the inner part of the sensor), and finally, in Step 6, an FFC connector was clinched to the connecting traces, using a 3D-printed tool designed to align the sensor leg contact leads with the connector. Details of the assembly process are depicted in Fig.~\ref{fig.sensorAssembly}. In total, 8 different Velostat sensors were created, one group utilizing a PET substrate together with 1, 2, 4, 6 layers of Velostat, and another group utilizing a TPU substrate together with 1, 2, 4, 6 layers of Velostat.  

\subsection{Printed Sensor Manufacturing}
\begin{figure}[!t]
\centering
\includegraphics[width=1\linewidth]{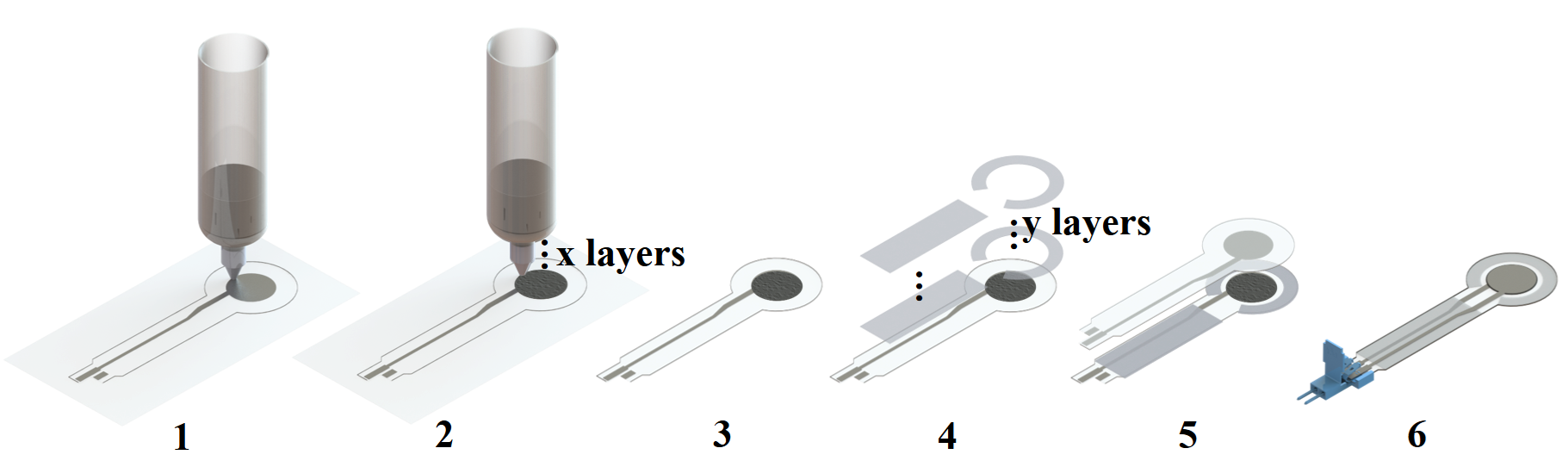}
\caption{Printed sensor detailed manufacturing process}
\label{fig.printedManufacturing}
\end{figure}

\begin{figure}[!t]
\centering
\subfloat[]{\includegraphics[width=.44\linewidth]{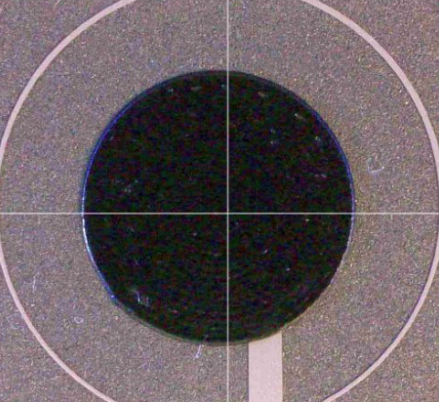}
\label{subfig.carbongood}}
\hfil
\subfloat[]{\includegraphics[width=.45\linewidth]{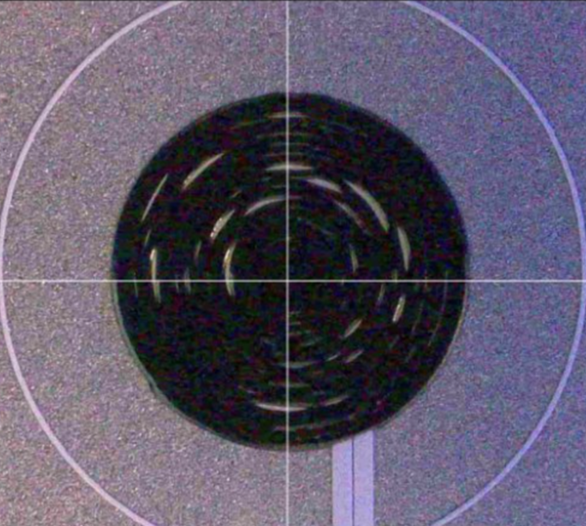}
\label{subfig.carbonbad}}
\caption{(a) Ideal printing demonstrating no spaces between traces (b) Poor printing with multiple spaces between traces}
\label{fig.carbonAll}
\end{figure}

The Printed sensor manufacturing process (Fig. \ref{fig.printedManufacturing}) started by loading the Gerber file for conductive, cutting, alignment, and calibration traces into the Voltera NOVA. Then a calibration process was performed for the conductive ink SS1109, the nozzle Nordson EFD 30GA (150 $\mu m$ diameter), and the substrate PET. Variables determined by the calibration process can be found in Table. \ref{tabInkSettings}. After calibration, the conductive and cutting traces were printed (Step 1) and the material was cured at 135°C for 5 minutes. After curing, the substrate was placed again in the Voltera NOVA and a secondary calibration process was performed for the carbon ink SC1502, the nozzle Nordson EFD 22GA (410 $\mu m$ diameter), and the substrate PET.
Variables determined by the calibration process can be found in Table.~\ref{tabInkSettings}.
Large-size nozzles (300-410 $\mu m$ diameter) are ideal for printing thick and wide layers. 
Thick traces allow for a thicker piezoresistive layer, creating a wider dynamic range of resistance values for the resulting sensor. Multiple piezoresistive layers are stacked through subsequent printing rounds to further increase the dynamic range.
After calibration, the sensing traces were printed and cured at 120°C for 15 minutes (Step 2). For each of x (1, 2, 3, or 4) sensing layers, Step 2 was repeated. After each printing process, the traces were evaluated before curing to make sure all of them had similar properties with good ink deposition. Ideally, a good ink deposition demonstrates a continuous path without gaps between traces, as shown in Fig. \ref{subfig.carbongood}, but even after calibrating the ink on the substrate, the ink deposition might still have defects (Fig. \ref{subfig.carbonbad}), which require a calibration fine-tuning. The final printing and dispenser settings are summarized in Table~\ref{tabInkSettings}. The Voltera NOVA reports the dispense-pressure and relief-pressure settings as internal controller values expressed in ADC units rather than as calibrated physical pressures. The dispense and relief thresholds are dimensionless ratios
reported as percentages.

After curing both the conductive and piezoresistive layers, the sensor is assembled. In Fig.~\ref{fig.printedManufacturing} Step 3, a scissor was used to cut along the cutting traces, which gave the sensor its final shape. Later, a y number of spacer layers were added (Step 4), followed by the addition of another layer of substrate with conductive and sensing layers printed, with the sensing layer side facing the inner side of the sensor (Step 5). Finally, in Step 6, an FFC connector was clinched to the connecting traces.
In total, 4 different printed sensors were created with 2, 4, 6, and 8 layers of printed material.
The assembly process depicted in Fig.~\ref{fig.sensorAssembly} for the Velostat sensor manufacturing is similar for the printed sensor manufacturing here, except that step (e) is removed for placing the Velostat layers and the prints in step (a) start with multiple layers of conductive and resistive traces.

\begin{table}[]
\begin{center}
\caption{Printing settings}
\label{tabInkSettings}
\begin{tabular}{|c|c|c|c|c|c|}
\hline
\textbf{Settings}                                                     & \textbf{SS1109}                                                & \textbf{SC1502}                                                & \textbf{Settings}                                                       & \textbf{SS1109}     & \textbf{SC1502}      \\ \hline
\begin{tabular}[c]{@{}c@{}}Print \\ Speed\end{tabular}       & \begin{tabular}[c]{@{}c@{}}200 \\ mm/min\end{tabular} & \begin{tabular}[c]{@{}c@{}}150 \\ mm/min\end{tabular} & \begin{tabular}[c]{@{}c@{}}Nozzle \\ Diameter\end{tabular}     & 150$\mu m$ & 410$\mu m$ \\ \hline
\begin{tabular}[c]{@{}c@{}}Trace \\ Space\end{tabular}       & 140$\mu m$                                            & 360$\mu m$                                            & \begin{tabular}[c]{@{}c@{}}Preheat \\ Temperature\end{tabular} & 30°C       & 30°C       \\ \hline
\begin{tabular}[c]{@{}c@{}}Print \\ Height\end{tabular}      & 65$\mu m$                                             & 170$\mu m$                                            & \begin{tabular}[c]{@{}c@{}}Dispense \\ Threshold\end{tabular}  & 5\%        & 0\%        \\ \hline
\begin{tabular}[c]{@{}c@{}}Dispense \\ Pressure\end{tabular} & 5900                                                  & 5900                                                  & \begin{tabular}[c]{@{}c@{}}Relief \\ Threshold\end{tabular}    & 0\%        & 0\%        \\ \hline
\begin{tabular}[c]{@{}c@{}}Relief \\ Pressure\end{tabular}   & 1800                                                   & 3000                                                  & \begin{tabular}[c]{@{}c@{}}Relief \\ Offset\end{tabular}       & 0.2mm      & 0.2mm       \\ \hline
\end{tabular}
\end{center}
\end{table}


\section{Experiments and Results}\label{sec.Experiments_and_Results}

\begin{figure}[!t]
\centering
\includegraphics[width=.75\linewidth]{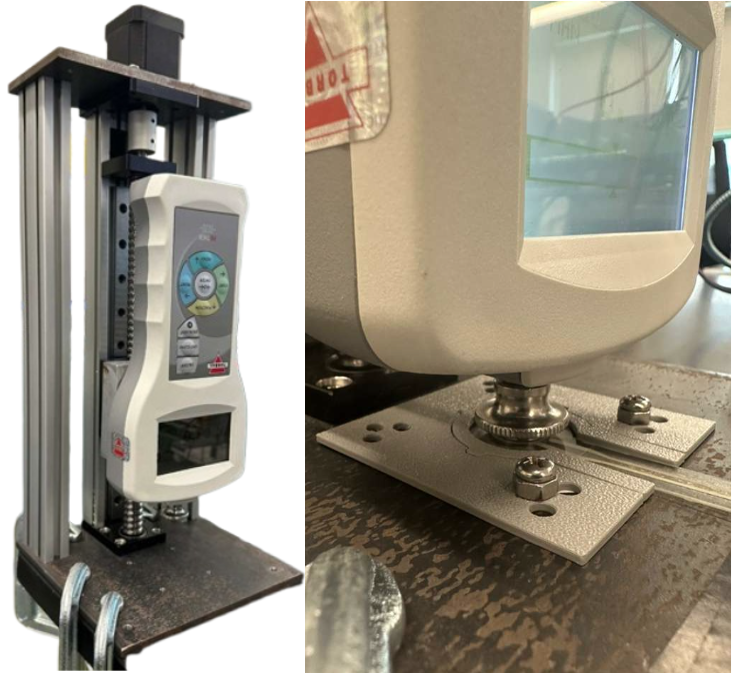}
\caption{(\emph{left}) Sensor testing machine, consisting of a motor-driven linear actuator controlling the downward motion of a Torbal FB200 force gauge.  (\emph{right}) Close-up of a sensor being compressed between the base steel plate and a steel puck attached to the head of the force gauge.}
\label{fig.torbal}
\end{figure}

All the manufactured sensors are evaluated with multiple metrics, including linearity, sensitivity, hysteresis, response time, drift, and cyclic stability. To evaluate the manufactured sensors, a machine equipped with a stepper motor-driven linear actuator and a force gauge (Torbal FB200) was utilized (Fig. \ref{fig.torbal}). This machine can perform both force and position control tests and can accurately measure forces with $0.4N$ accuracy and $0.05N$ resolution. 
For drift experiments, closed-loop force control was implemented using force feedback from the Torbal.
For all other experiments, open-loop positions control was used.
Steel pucks were attached to the Torbal head for contact with the sensors during the compressive actuation. These pucks were machined to have areas equal to 70-85\% of the sensor's sensing area.


\subsection{Data Collection}

\subsubsection{Hysteresis, Linearity, and Sensitivity}

For the hysteresis, linearity, and sensitivity tests, each sensor was
loaded to a selected maximum force using a staircase triangular loading
profile under position control. Each displacement step was maintained for
0.2~s during both loading and unloading, and the stepper motor was operated
at 2000 steps/revolution. The resulting force--output curves were used to
quantify sensitivity, linearity, and hysteresis.

Because the sensor responses were strongly nonlinear, sensitivity was
evaluated as a function of force rather than represented by a single
full-range value. The
local sensitivity was then calculated as

\begin{equation}
S(F)=\frac{dV}{dF},
\label{eq:sensitivity}
\end{equation}

where $V$ is the measured sensor output voltage and $F$ is the applied
force.

Linearity was evaluated by fitting the loading response over the selected
analysis range with a best-fit straight line,

\begin{equation}
\hat{V}(F)=aF+b,
\label{eq:linear_fit}
\end{equation}

where $a$ and $b$ are the least-squares slope and intercept, respectively.
The coefficient of determination was calculated as

\begin{equation}
R^2
=
1-
\frac{
\sum_{i=1}^{N}\left[V_i-\hat{V}_i\right]^2
}{
\sum_{i=1}^{N}\left[V_i-\overline{V}\right]^2
},
\label{eq:r_squared}
\end{equation}

where $V_i$ is the measured output, $\hat{V}_i$ is the corresponding
value predicted by the best-fit straight line, and $\overline{V}$ is the mean measured output. 

Hysteresis was evaluated by comparing the loading and unloading responses
at identical applied forces. Because the experimentally recorded loading
and unloading data were not sampled at exactly the same force values,
both curves were linearly interpolated onto a common force vector over
their overlapping force range. To characterize hysteresis over the complete analyzed force range rather than only at the point of maximum separation, a normalized hysteresis-area metric was also calculated as

\begin{equation}
H_{\mathrm{area}}
=
\frac{
\displaystyle
\int_{F_{\min}}^{F_{\max}}
\left|
V_{\mathrm{load}}(F)-V_{\mathrm{unload}}(F)
\right|\,dF
}{
\left(F_{\max}-F_{\min}\right)V_{\mathrm{FS}}
}
\times 100\%.
\label{eq:area_hysteresis}
\end{equation}

The numerator of (\ref{eq:area_hysteresis}) represents the area between
the loading and unloading curves, while the denominator normalizes this
area by the rectangular area defined by the analyzed force range and
full-scale sensor output.

\subsubsection{Response Time}
\begin{figure}[!t]
\centering
\includegraphics[width=0.75\linewidth]{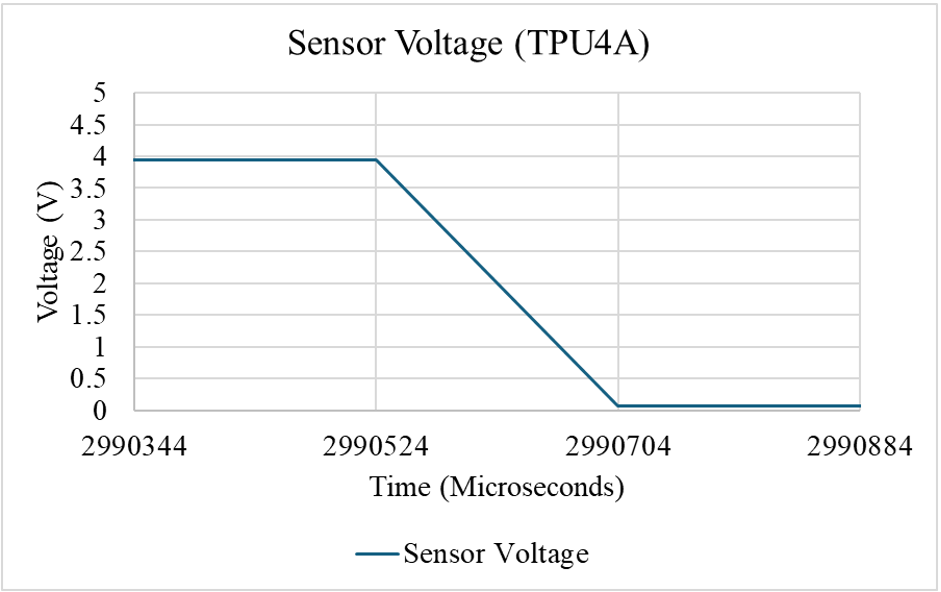}
\caption{Response time test example}
\label{fig.ResponseTime}
\end{figure}

In the response-time test (Fig. \ref{fig.ResponseTime}), the sensor was initially loaded to a force between 10--20~N under position control and then rapidly unloaded. The unloading motion was performed using the lowest stepper-motor setting of 200 steps/revolution and the maximum motor speed. The test was intended to determine the time required for the sensor output to reach 63.2\% of the total signal change following unloading~\cite{nie2024mechanically}. However, the measured output transitions occurred within the temporal resolution of the acquisition system, which was $200~\mu\mathrm{s}$, preventing the intrinsic sensor response time from being resolved. Therefore, response time could not be quantitatively determined using the present experimental apparatus and is reported as a limitation of this study.

\subsubsection{Drift}
\begin{figure}[!t]
\centering
\includegraphics[width=0.75\linewidth]{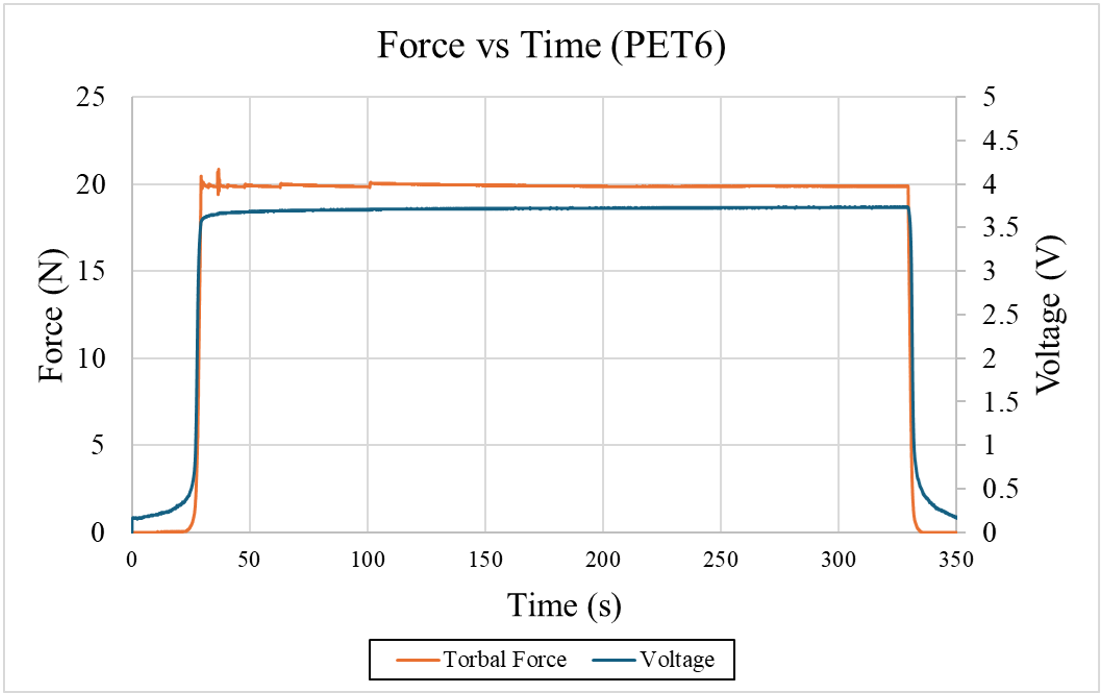}
\caption{Drift test example}
\label{fig.Drift}
\end{figure}

In the Drift Test, the sensor was stressed by a 20N force. This force was maintained for a total of 5 minutes using force control to estimate how much the sensor voltage varies per minute at a constant load. An example of a drift test for one of the sensors, PET6, is shown in Fig. \ref{fig.Drift}.


\subsubsection{Cyclic Stability}
In this test, the sensor was stressed by a 10N force triangle wave with 1/3 Hz frequency for a total of 10,000 Cycles. This test was used to understand if the sensor output is repeatable or if it substantially varies after a specific number of iterations.


\subsection{Signal Acquisition}
\label{sec:signal_acquisition}

\begin{figure}[!t]
    \centering
    \includegraphics[width=0.55\columnwidth]
    {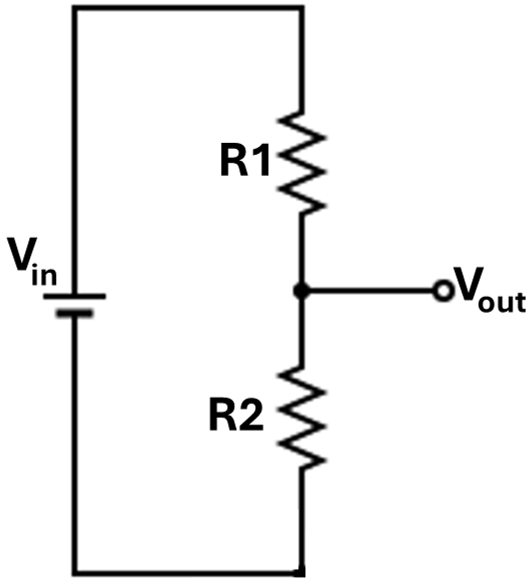}
    \caption{Voltage-divider circuit used for sensor signal acquisition.}
    \label{fig:circuitVoltageDivider}
\end{figure}

All sensors were evaluated using the voltage-divider circuit shown in
Fig.~\ref{fig:circuitVoltageDivider}. The force-sensitive resistance,
$R_{\mathrm{FSR}}$, was connected in series with a fixed bias resistor,
$R_b$, and supplied with a known input voltage, $V_{\mathrm{in}}$.
Because the unloaded and loaded resistance values differed among the
sensor configurations, the bias resistor was selected separately for
each sensor.

To select the bias resistor, each sensor was loaded between its
preselected minimum and maximum test forces while its resistance was
measured. A bias resistor was then selected to use as much as practical
of the available 0--5~V acquisition range without saturating the output.
The selected bias resistors are listed in Table~\ref{tab:bias_resistors}.
Once selected, the same bias resistor was used for all characterization
experiments performed on the corresponding sensor.

\begin{table}[!t]
    \caption{Selected Bias Resistor for Each Sensor}
    \label{tab:bias_resistors}
    \centering
    \begin{tabular}{|c|c|c|c|}
        \hline
        \textbf{Sensor} & \textbf{Bias Resistor} &
        \textbf{Sensor} & \textbf{Bias Resistor} \\
        \hline
        PET1   & $68\,\Omega$  & TPU1 & $47\,\Omega$  \\
        \hline
        PET2   & $150\,\Omega$ & TPU2 & $68\,\Omega$  \\
        \hline
        PET4   & $330\,\Omega$ & TPU4 & $220\,\Omega$ \\
        \hline
        PET6   & $220\,\Omega$ & TPU6 & $470\,\Omega$ \\
        \hline
        Tangio & $5600\,\Omega$  & Tekscan & $200K\,\Omega$  \\
        \hline
    \end{tabular}
\end{table}

For the voltage-divider configuration used in this study, the output
voltage is given by

\begin{equation}
    V_{\mathrm{out}}
    =
    V_{\mathrm{in}}
    \left(
        \frac{R_b}
        {R_b + R_{\mathrm{FSR}}}
    \right),
    \label{eq:voltage_divider}
\end{equation}

where $V_{\mathrm{out}}$ is the measured output voltage,
$V_{\mathrm{in}}$ is the input voltage, $R_b$ is the fixed bias
resistance, and $R_{\mathrm{FSR}}$ is the instantaneous resistance of
the force sensor.

The sensor resistance can therefore be calculated from the measured
output voltage as

\begin{equation}
    R_{\mathrm{FSR}}
    =
    R_b
    \left(
        \frac{V_{\mathrm{in}}}{V_{\mathrm{out}}}
        - 1
    \right).
    \label{eq:sensor_resistance}
\end{equation}

Equation~\eqref{eq:sensor_resistance} allows the force-dependent
resistance of each sensor to be determined from the measured output
voltage and the corresponding bias-resistor value.

For comparison of resistance changes among sensors, the normalized resistance change was calculated as

\begin{equation}
\frac{\Delta R}{R_0}
=
\frac{R_0-R}{R_0},
\end{equation}

where $R_0$ is the unloaded baseline resistance and $R$ is the
instantaneous sensor resistance.

\subsection{Results}
\label{sec:results}

All results reported in this section were obtained using the
voltage-divider readout described in Section~\ref{sec:signal_acquisition}.

\begin{figure}[!t]
\centering
\subfloat[]{\includegraphics[width=.45\linewidth]{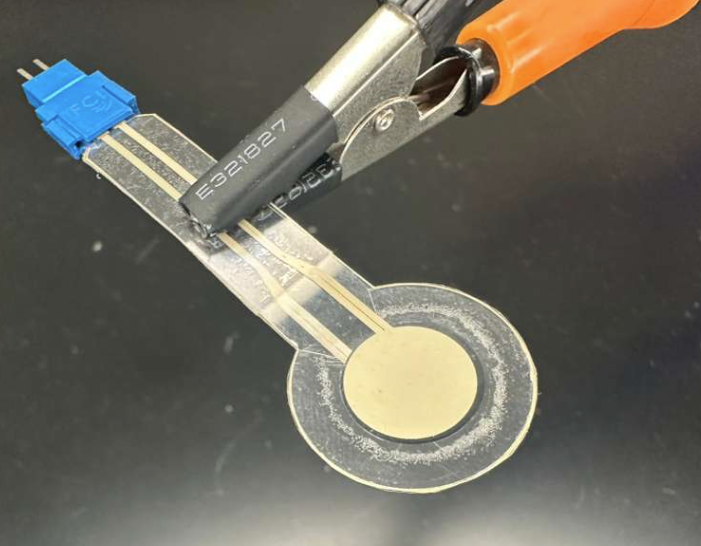}
\label{subfig.sensorStiff}}
\hfil
\subfloat[]{\includegraphics[width=.45\linewidth]{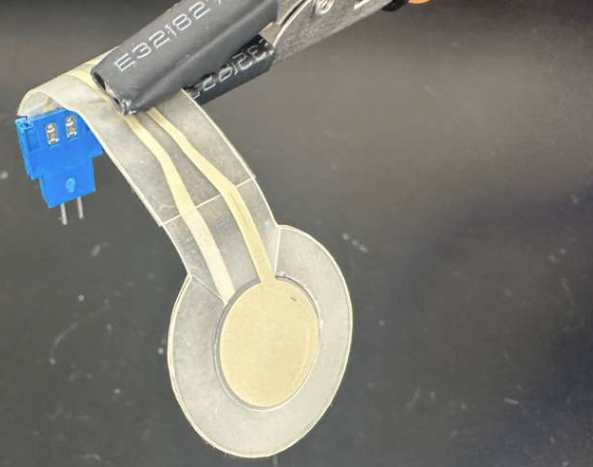}
\label{subfig.sensorStretch}}
\caption{(a) Stiff sensor created with PET (b) Conformable sensor created with TPU}
\label{fig:sensorTPUPET}
\end{figure}

The fabrication workflow produced eight multilayer Velostat
force-sensor configurations containing 1, 2, 4, and 6 Velostat
layers on both PET and TPU substrates. Representative completed
sensors are shown in Fig.~\ref{fig:sensorTPUPET}. The PET-based sensor retained a
comparatively fixed planar form, whereas the TPU-based sensor
could conform more readily to a curved surface. The force-dependent performance of these configurations is compared
quantitatively in the following subsections.

Four printed-carbon sensor configurations containing 2, 4, 6,
and 8 printed sensing layers were also fabricated. Because the
printed-carbon architecture was included as an exploratory
demonstration of a more fully printed sensor fabricated using the
Voltera NOVA, its fabrication outcome and force-dependent response
are discussed separately in Section~\ref{sec:printedCarbon}.

\subsubsection{Hysteresis, Linearity, and Sensitivity}
\begin{figure*}
\centering
\includegraphics[width=0.95\linewidth]{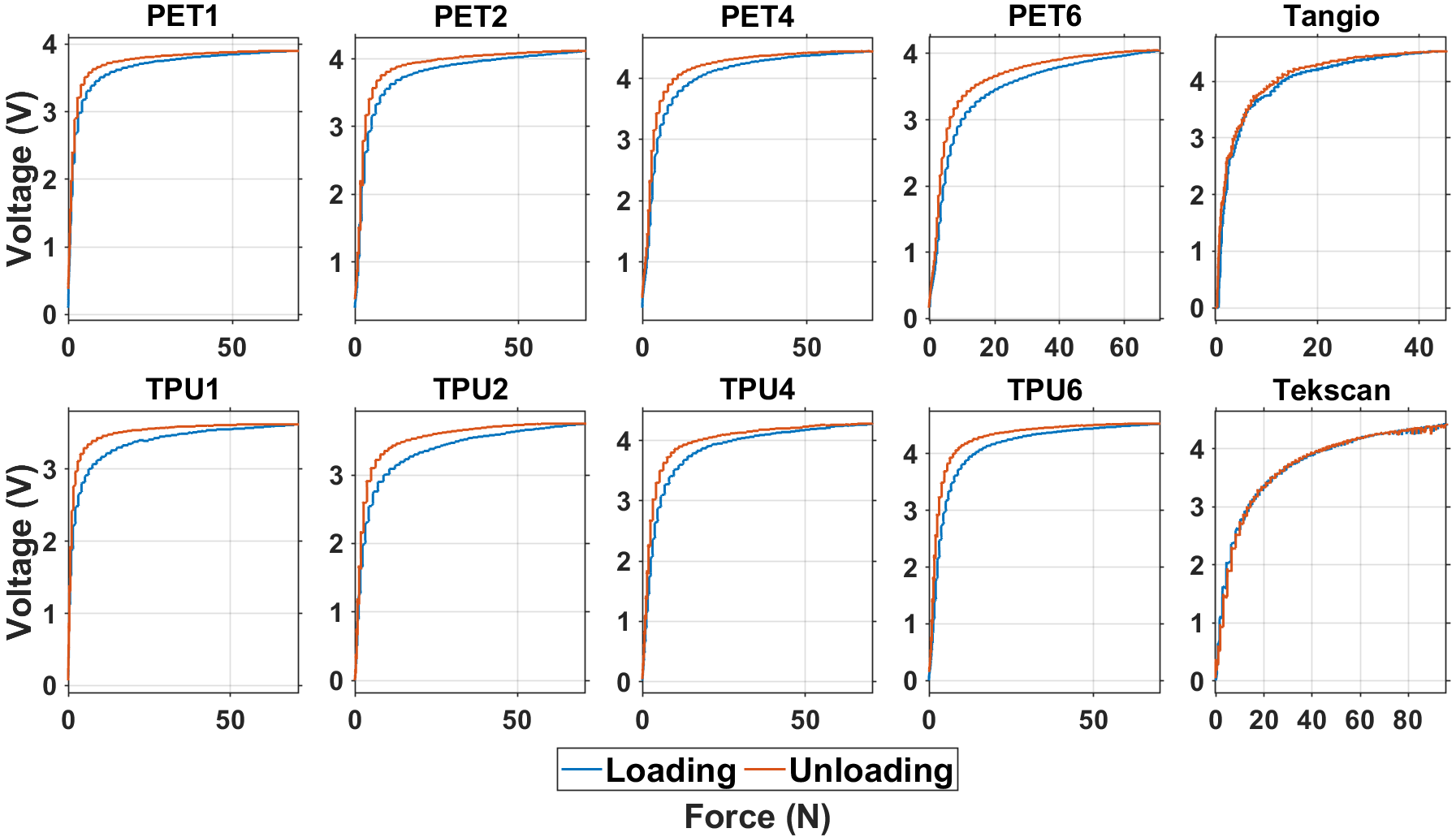}
\caption{Measured force--voltage responses during loading and unloading for the fabricated Velostat sensors and the commercial Tangio and Tekscan sensors.}
\label{fig.hysteresisFvsVPlot}
\end{figure*}

Figure~\ref{fig.hysteresisFvsVPlot} shows the measured force--voltage responses during loading and unloading for the fabricated Velostat sensors and the commercial Tangio and Tekscan sensors. All tested Velostat sensors exhibited strongly nonlinear responses, characterized by a rapid increase in output at low forces followed by a progressively reduced slope at higher forces. This behavior is reflected in the best-fit straight-line (BFSL) $R^2$ values reported in Table~\ref{tab:linearity_hysteresis}, which ranged from 0.475 for PET2 to 0.641 for PET6. The commercial Tangio and Tekscan sensors exhibited somewhat higher $R^2$ values of 0.656 and 0.706, respectively, although their responses also remained nonlinear. No consistent monotonic relationship between Velostat layer count and BFSL $R^2$ was observed.

\begin{figure*}
\centering
\includegraphics[width=0.95\linewidth]{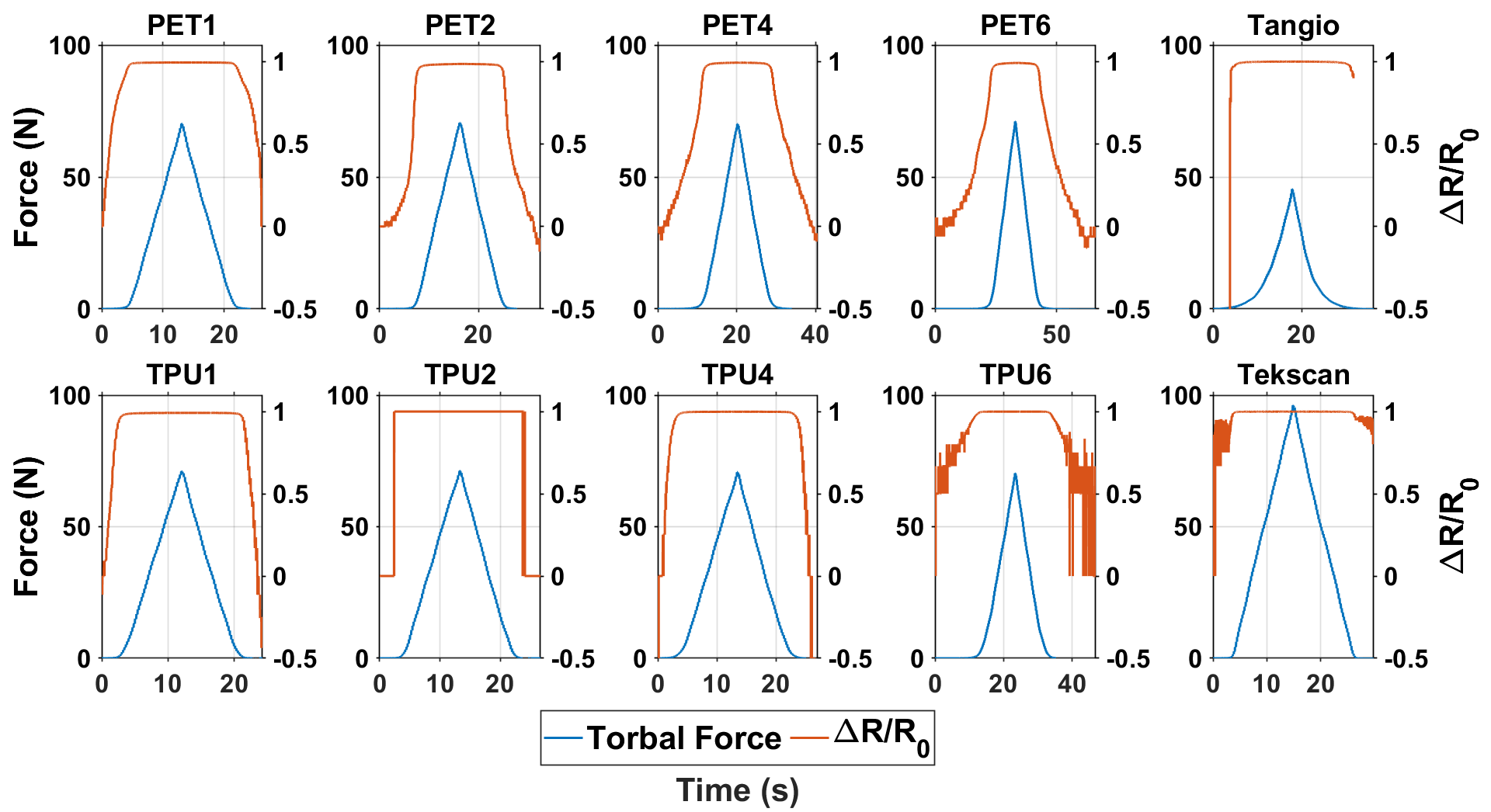}
\caption{Applied force and normalized resistance change, $\Delta R/R_0$, as functions of time during loading and unloading for the fabricated Velostat sensors and the commercial Tangio and Tekscan sensors.}
\label{fig.hysteresisTimePlot}
\end{figure*}

Hysteresis was quantified using the normalized area between the loading and unloading curves. As summarized in Table~\ref{tab:linearity_hysteresis}, the Velostat sensors exhibited area-based hysteresis values ranging from 3.90\% to 7.47\%. PET4 exhibited the lowest hysteresis among the tested Velostat sensors at 3.90\%, whereas TPU1 exhibited the highest at 7.47\%. The Tangio and Tekscan sensors exhibited lower hysteresis values of 2.50\% and 0.77\%, respectively. Similar to the linearity results, no monotonic relationship between Velostat layer count and hysteresis was observed. Figure~\ref{fig.hysteresisTimePlot} additionally shows the applied force and normalized resistance change, $\Delta R/R_0$, as functions of time during the loading--unloading cycle.

\begin{figure}[!t]
\centering
\includegraphics[width=0.95\linewidth]{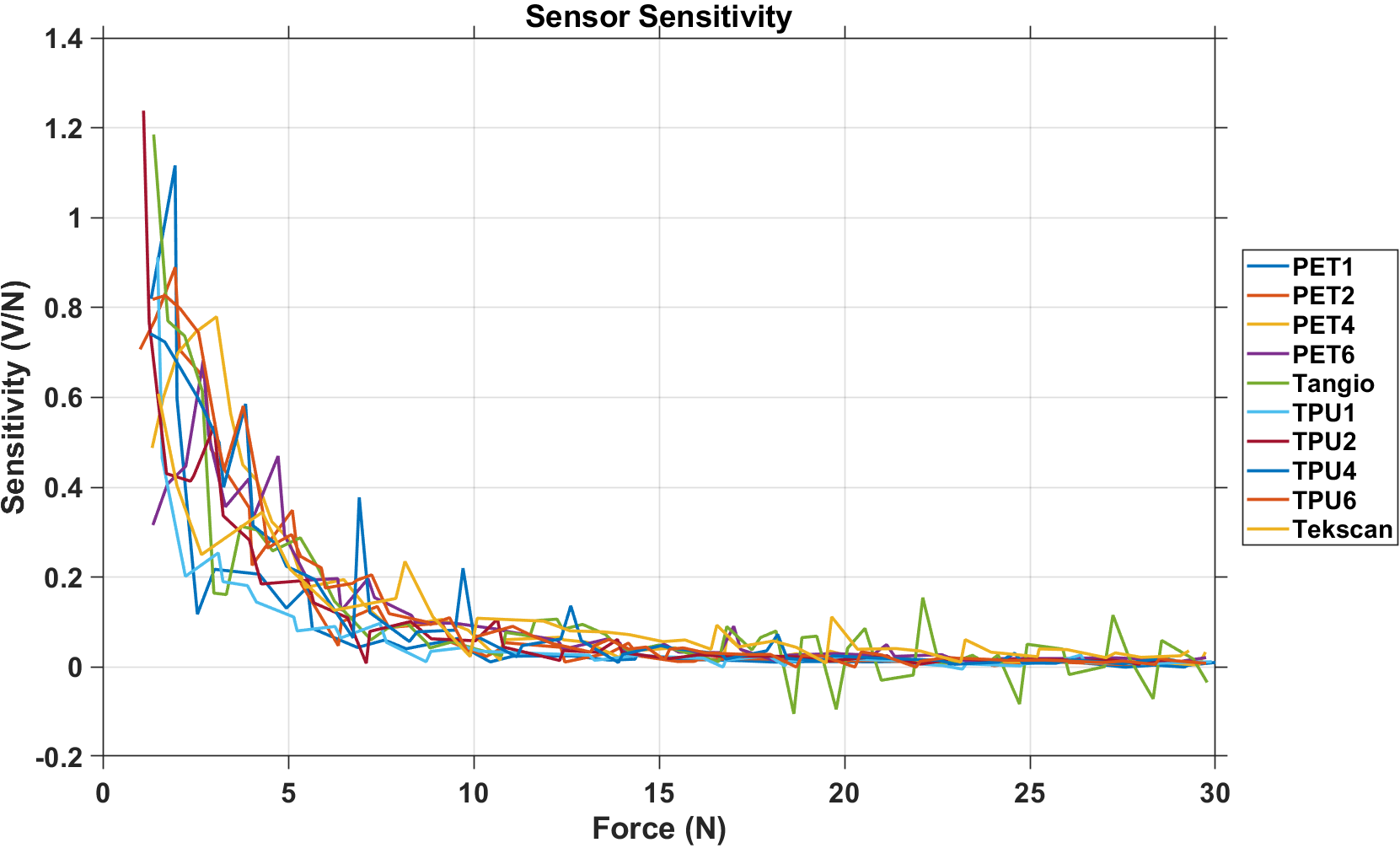}
\caption{Force-dependent sensitivity of the fabricated Velostat sensors and the commercial Tangio and Tekscan sensors. Sensitivity was calculated as the local slope of the force--voltage response, $S=dV/dF$, illustrating the higher sensitivity at low applied forces and the progressive decrease in sensitivity with increasing force.}
\label{fig.SensorSensitivityComparison}
\end{figure}

Because the force--voltage responses were strongly nonlinear, a single sensitivity value was not considered representative of the complete operating range. Figure~~\ref{fig.SensorSensitivityComparison} therefore presents the local sensitivity, calculated as $S=dV/dF$, as a function of applied force. Sensitivity was greatest at low forces, with some sensors reaching values greater than 1~V/N, and decreased rapidly as force increased. For most sensors, the sensitivity approached approximately 0--0.2~V/N at higher forces. This reduction in sensitivity with increasing force is consistent with the saturating behavior observed in Fig.~\ref{fig.hysteresisFvsVPlot} and indicates that the tested sensors are substantially more sensitive in the lower portion of their force ranges.

\begin{table}[!t]
    \caption{Quantitative Linearity and Hysteresis Results}
    \label{tab:linearity_hysteresis}
    \centering
    \renewcommand{\arraystretch}{1.05}
    \begin{tabular}{|c|c|c|c|c|c|}
        \hline
        \textbf{Sensor} &
        $\boldsymbol{R^2}$ &
        $\boldsymbol{H_{\mathrm{area}}}$ \textbf{(\%)} &
        \textbf{Sensor} &
        $\boldsymbol{R^2}$ &
        $\boldsymbol{H_{\mathrm{area}}}$ \textbf{(\%)} \\
        \hline
        PET1   & 0.48933 & 3.9406 &
        TPU1   & 0.61708 & 7.4678 \\
        \hline
        PET2   & 0.47514 & 3.9753 &
        TPU2   & 0.60512 & 6.3509 \\
        \hline
        PET4   & 0.51091 & 3.8982 &
        TPU4   & 0.55579 & 4.2798 \\
        \hline
        PET6   & 0.64063 & 5.0111 &
        TPU6   & 0.49993 & 4.6060 \\
        \hline
        Tangio & 0.65565 & 2.5017 &
        Tekscan & 0.70602 & 0.7714 \\
        \hline
    \end{tabular}
\end{table}




\subsubsection{Drift}
\label{sec:drift_results}
\normalcolor
Table~\ref{tab:voltage_drift} summarizes the change in the
voltage-divider output while each sensor was maintained at a constant
force of $20~\mathrm{N}$ for $5~\mathrm{min}$. Here, $V_0$ is the
output voltage measured immediately after the target force was reached,
and $V_{5\mathrm{min}}$ is the output voltage measured after
$5~\mathrm{min}$ under the constant force. The percentage voltage drift
was calculated as

\begin{equation}
    D_V
    =
    \frac{
        \left|V_{5\mathrm{min}}-V_0\right|
    }{
        V_0
    }
    \times 100\%.
    \label{eq:voltage_drift}
\end{equation}

Among the tested devices, the measured voltage drift ranged from 1.9\% to
7.3\%. TPU2 exhibited the greatest voltage drift at 7.3\%, followed by TPU4
at 5.4\%. The remaining Velostat configurations exhibited voltage drift below
5\%. The commercial Tangio and Tekscan sensors exhibited drift values of
3.0\% and 2.96\%, respectively. No consistent monotonic relationship between
the number of Velostat layers and voltage drift was observed.


\begin{table}[!t]
    \caption{Voltage Drift Under a Constant Applied Force}
    \label{tab:voltage_drift}
    \centering
    \renewcommand{\arraystretch}{1.05}
    \begin{tabular}{|c|c|c|c|c|}
        \hline
        \textbf{Sensor} &
        $\boldsymbol{V_0}$ \textbf{(V)} &
        $\boldsymbol{V_{5\mathrm{min}}}$ \textbf{(V)} &
        $\boldsymbol{\Delta V}$ \textbf{(V)} &
        $\boldsymbol{D_V}$ \textbf{(\%)} \\
        \hline
        PET1   & 3.64 & 3.75 & 0.11 & 3.0 \\
        \hline
        PET2   & 3.86 & 3.99 & 0.13 & 3.4 \\
        \hline
        PET4   & 3.88 & 4.06 & 0.18 & 4.6 \\
        \hline
        PET6   & 3.56 & 3.73 & 0.17 & 4.8 \\
        \hline
        Tangio & 4.24 & 4.37 & 0.13 & 3.0\\
        \hline
        TPU1   & 3.66 & 3.73 & 0.07 & 1.9 \\
        \hline
        TPU2   & 3.15 & 3.38 & 0.23 & 7.3 \\
        \hline
        TPU4   & 3.67 & 3.87 & 0.20 & 5.4 \\
        \hline
        TPU6   & 4.16 & 4.30 & 0.14 & 3.4 \\
        \hline
        Tekscan & 3.38 & 3.48 & 0.1 & 2.96 \\
        \hline
    \end{tabular}
\end{table}



\subsubsection{Cyclic Stability}
\label{sec:Cyclic Stability}

\begin{figure}[!t]
\centering
\subfloat[]{\includegraphics[width=1\linewidth]{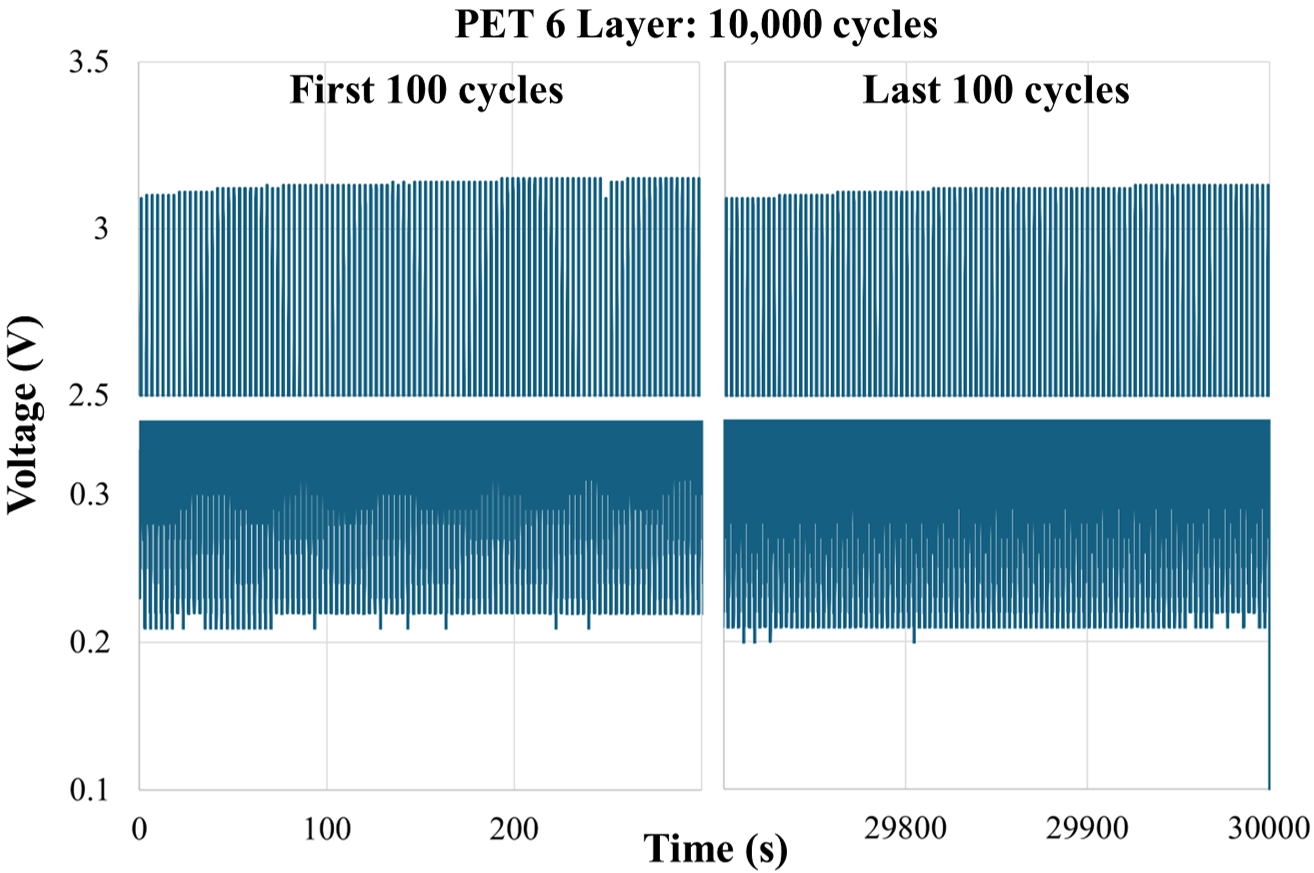}
\label{subfig.repeatabilityPet6}}
\hfil
\subfloat[]{\includegraphics[width=1\linewidth]{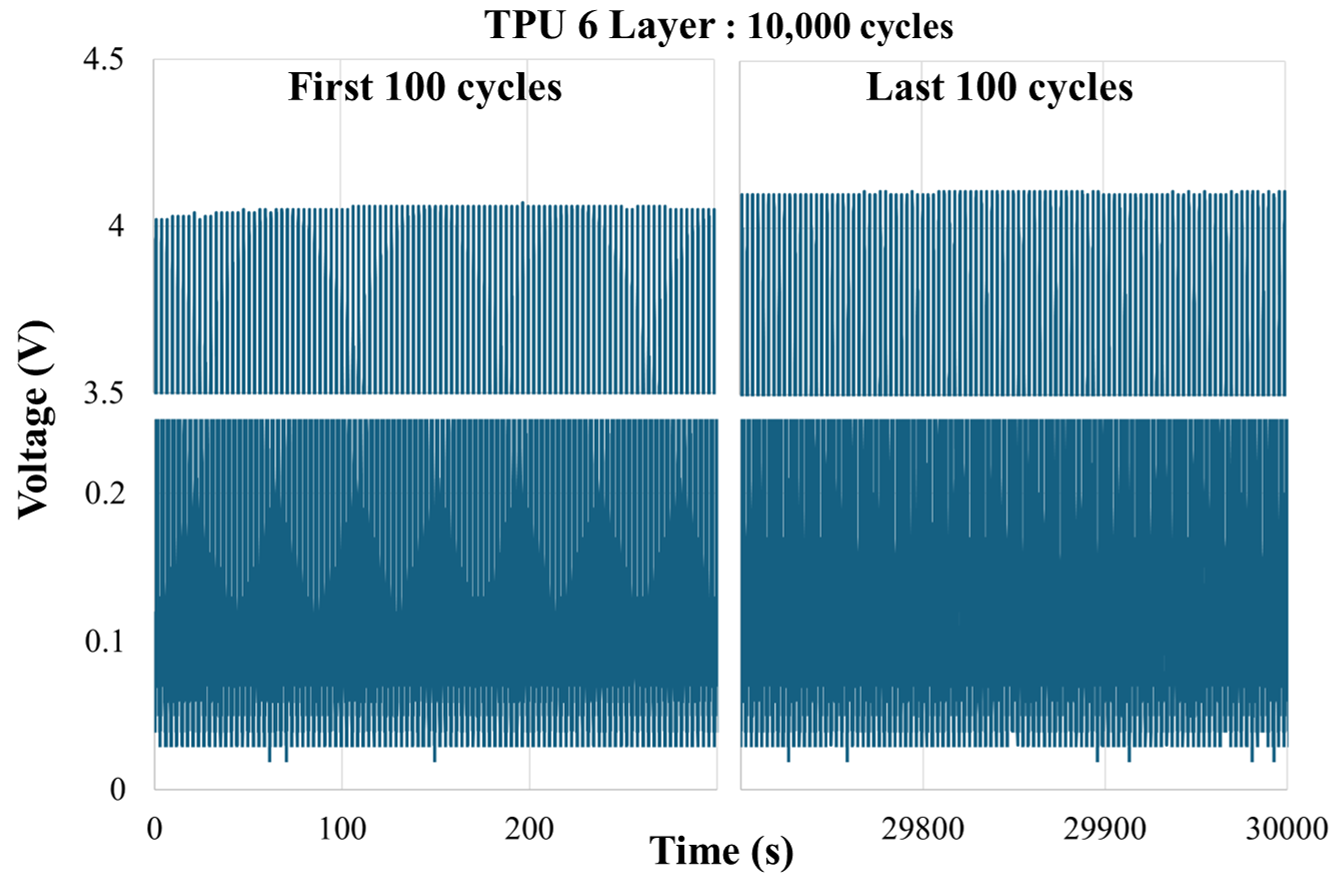}
\label{subfig.repeatabilityTpu6}}
\hfil
\caption{Cyclic stability tests for (a) PET 6-layer, (b) TPU 6-layer. For each of (a),(b): \emph{left} shows first 100 cycles, and \emph{right} shows last 100 cycles, out of a total of 10,000 actuation cycles.}
\label{fig.repeatabilityAll}
\end{figure}

The PET6, TPU6, and 8-layer printed-carbon sensors were
subjected to 10,000 loading--unloading cycles between the
unloaded condition and an applied force of $10~\mathrm{N}$ at a
frequency of $1/3~\mathrm{Hz}$. Figure~\ref{fig.repeatabilityAll}
compares the voltage-divider outputs recorded during the first 100 and last 100 cycles of each test.

The PET6 and TPU6 sensors maintained similar peak output voltages
during the first and final 100-cycle intervals. Based on the
measured data, the change in peak output voltage between these
intervals was less than $0.05~\mathrm{V}$ for both sensors. Each
sensor also continued to return approximately to its unloaded
baseline after each loading cycle. These observations indicate
limited endpoint degradation of the PET6 and TPU6 sensor outputs
over 10,000 cycles.

\subsubsection{Exploratory Printed-Carbon Sensor}
\label{sec:printedCarbon}
The printed-carbon sensor was included as an exploratory demonstration of a
more fully printed sensor architecture in which both the conductive electrodes
and sensing element were fabricated using the Voltera NOVA. Unlike the
Velostat sensors, which incorporated a discrete piezoresistive sensing layer,
the printed-carbon sensor used ACI SC1502 carbon ink as the sensing material.
Of the printed-carbon configurations fabricated, the 2-, 4-, and 6-layer
devices either saturated below approximately 0.5~N or did not provide a usable
output; therefore, only the 8-layer configuration was characterized further.

\begin{figure}[!t]
    \centering
    \includegraphics[width=1.0\linewidth]
    {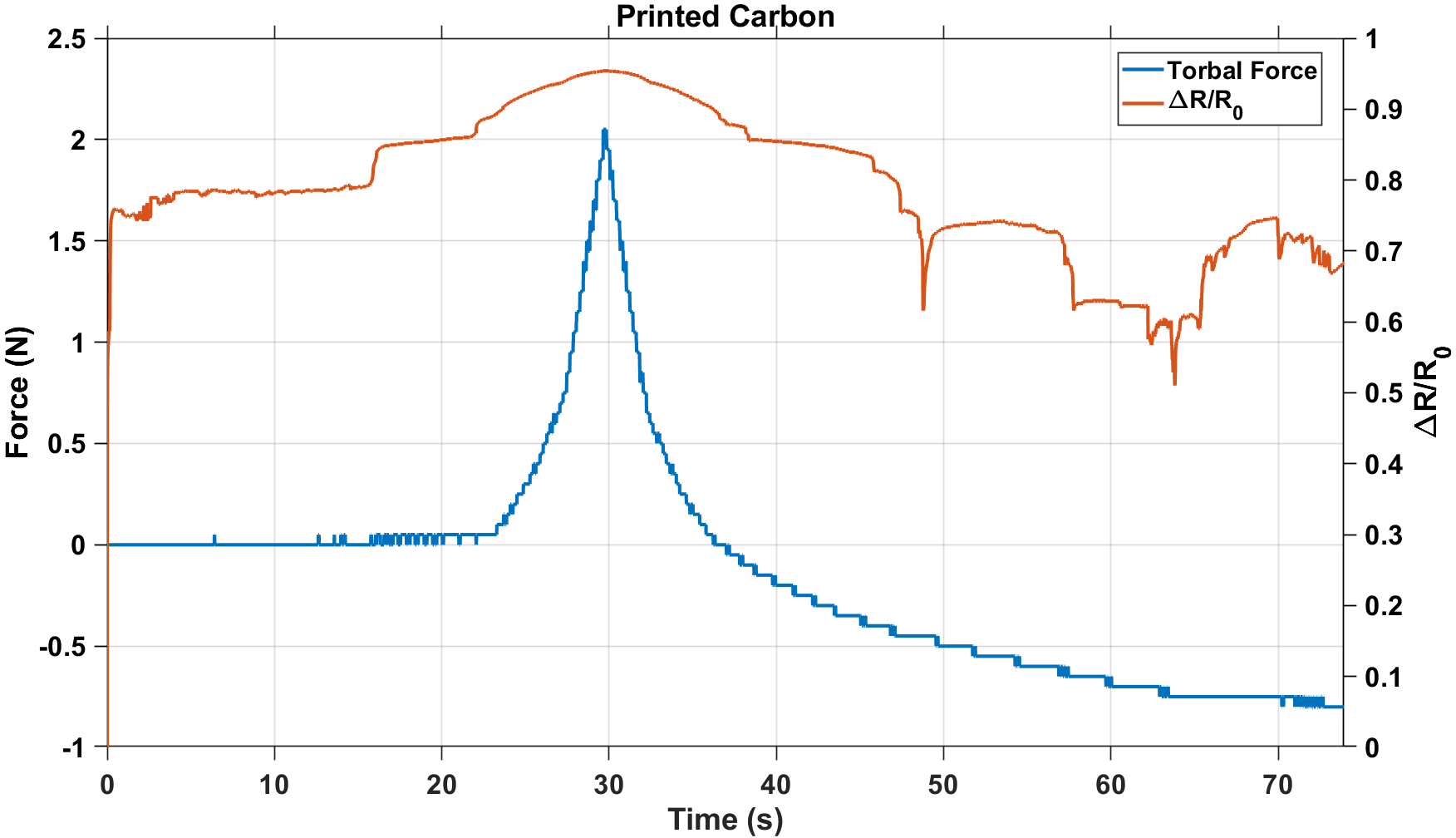}
    \caption{Exploratory printed-carbon sensor response showing the applied
    force and normalized resistance change, $\Delta R/R_0$, as functions of
    time.}
    \label{fig.printedCarbonResponse}
\end{figure}

As shown in Fig.~\ref{fig.printedCarbonResponse}, the 8-layer printed-carbon
sensor produced a measurable force-dependent response, but its useful operating
range was limited to approximately 2~N and the output exhibited substantial
variation. A likely contributor to this behavior is the use of ACI SC1502,
which is formulated as a stretchable carbon conductor for printed circuitry
rather than as a purpose-designed piezoresistive force-sensing ink. In addition,
curing-related shrinkage and deformation of the printed carbon layers were
observed during fabrication. These results demonstrate the feasibility of
fabricating the complete sensor architecture using the Voltera NOVA, while
indicating that further optimization of the sensing material, printed-layer
thickness, and curing conditions is required to obtain a reliable force-sensing response.




\section{Discussion}
The fabricated Velostat sensors exhibited strongly nonlinear force--voltage
responses, with the greatest sensitivity occurring at low forces and a
progressive decrease in sensitivity as force increased. The measured BFSL
$R^2$ values ranged from 0.475 to 0.641 and the area-based hysteresis ranged
from 3.90\% to 7.47\%. No consistent monotonic relationship was observed
between Velostat layer count and linearity, hysteresis, or drift, indicating
that layer count alone did not determine performance within the tested
architecture. 

The commercial Tangio and Tekscan sensors exhibited higher $R^2$ values and
lower hysteresis than the fabricated Velostat sensors, although their voltage
drift values were within the range measured for the Velostat devices. The PET
and TPU substrates also provided different mechanical characteristics: PET
produced a comparatively stiff planar sensor, whereas TPU provided greater
conformability. Because no consistent electrical-performance advantage was
observed for either substrate, substrate selection may instead depend on the
mechanical requirements of the intended application.

The PET6 and TPU6 devices showed limited endpoint degradation over 10,000
loading cycles, with changes in peak voltage below 0.05~V between the first
and final 100-cycle intervals. However, these results characterize only the
individual tested devices and do not establish device-to-device repeatability
or manufacturing reproducibility. 

The printed-carbon architecture should be considered exploratory. Although the
eight-layer device produced a measurable response, its useful force range was
limited to approximately 2~N, while configurations with fewer layers either
saturated below approximately 0.5~N or did not provide a usable response.
A likely contributor is the use of ACI SC1502, which is intended as a
stretchable carbon conductor rather than a purpose-designed piezoresistive
sensing ink. Further optimization of sensing material, printed-layer
thickness, curing conditions, and preload is therefore required. 

\section{Conclusion}\label{sec.Conclusion}
\normalcolor
This work developed and experimentally evaluated a parameterized
direct-ink-writing workflow for fabricating flexible force sensors using
printed silver electrodes with multilayer Velostat sensing elements on PET
and TPU substrates. Functional Velostat sensors were successfully fabricated
with 1, 2, 4, and 6 sensing layers. The tested devices exhibited nonlinear,
force-dependent sensitivity, measurable hysteresis and drift, and no
consistent monotonic relationship between Velostat layer count and the
measured performance metrics. Selected PET6 and TPU6 sensors also maintained
similar outputs over 10,000 loading cycles. 

An exploratory printed-carbon architecture was also fabricated entirely using
printed conductive and sensing layers with the Voltera NOVA. Although the
eight-layer device produced a measurable force-dependent response, the
SC1502-based architecture did not provide sufficiently reliable performance
for use as a force sensor in its present form. Overall, the results establish
the feasibility of the proposed printed-electrode Velostat fabrication
workflow while identifying the printed sensing material as a key area for further development.

Future work will evaluate multiple independently fabricated sensors for each
configuration to quantify device-to-device variability and manufacturing
reproducibility. Faster acquisition hardware will be used to resolve sensor
response time. For the
printed-carbon architecture, future studies will investigate sensing materials
specifically designed for piezoresistive force sensing and optimize printed
layer thickness, curing conditions, spacer geometry, and preload. A lower-range
force measurement system will also be used to characterize the printed-carbon
devices within their intended operating range.

\bibliographystyle{IEEEtran}
\bibliography{references}

\begin{IEEEbiography}[{\includegraphics[width=1in,height=1.25in,clip,keepaspectratio]{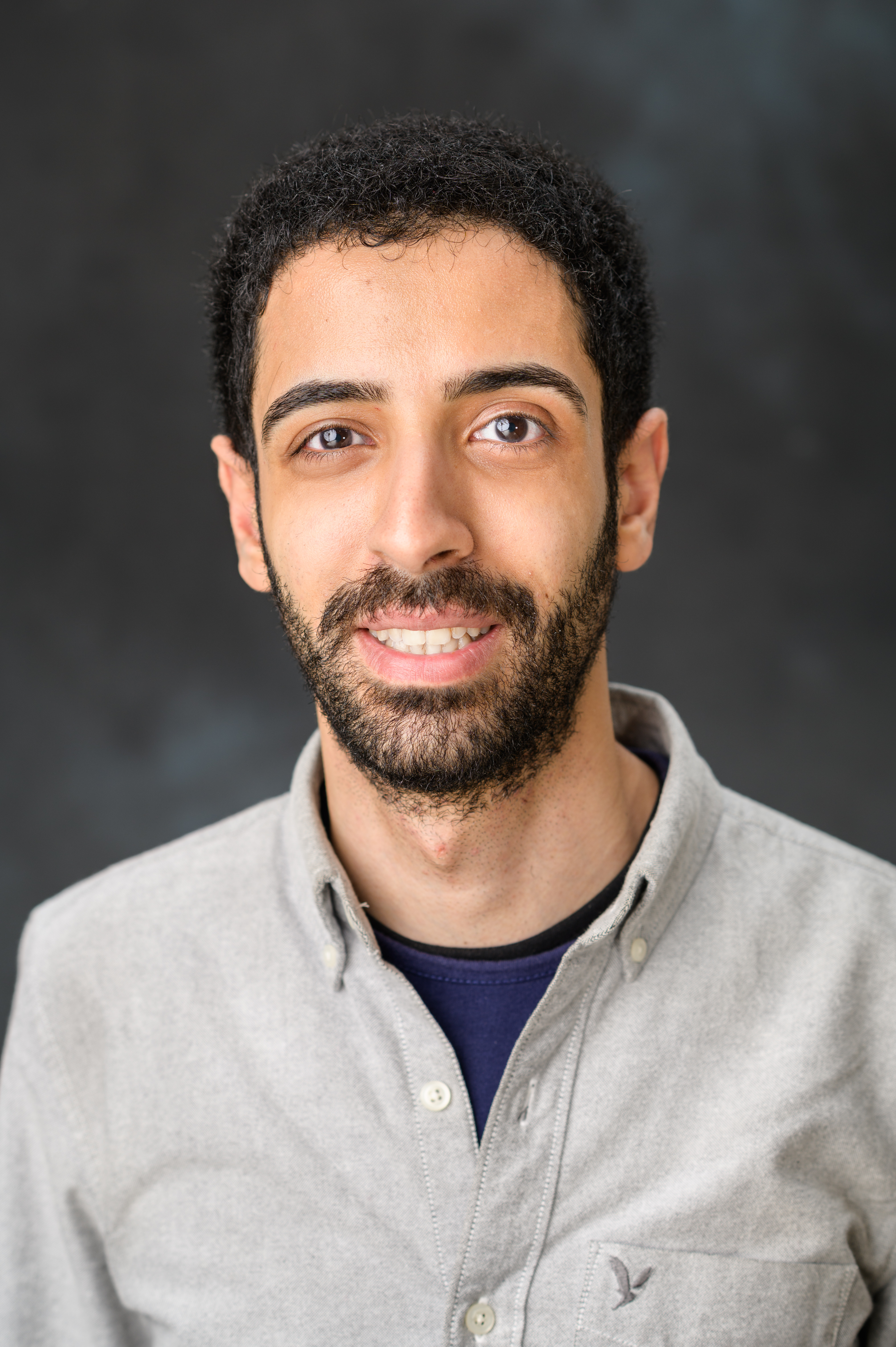}}]{Mostafa Mo. Massoud} (Graduate Student Member, IEEE) received the B.S. and M.S. degrees in 2017 and 2022, respectively, both in mechanical engineering with a focus on mechatronics from the Arab Academy for Science and Technology, Cairo, Egypt. He is working toward a Ph.D. in the Robotic Systems for Health Laboratory, advised by Professor Jacqueline Libby, in the Mechanical Engineering Department with a focus on mechatronics, controls, and soft robotics at Stevens Institute of Technology, Hoboken, NJ, USA.
\end{IEEEbiography}

\begin{IEEEbiography}[{\includegraphics[width=1in,height=1.25in,clip,keepaspectratio]{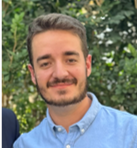}}]{Paulo H.T.F. Alves} (Graduate Student Member, IEEE) received the B.S. in Mechanical Engineering from Federal University of Ouro Preto, Ouro Preto, MG, Brazil.  He received the M.Sc. in Mechanical Engineering in 2023 from the University of Minnesota Duluth, Duluth, MN, USA. He is working toward a Ph.D. in the Robotic Systems for Health Laboratory, advised by Professor Jacqueline Libby, in the Mechanical Engineering Department at Stevens Institute of Technology, Hoboken, NJ, USA.  His research focuses on additive manufacturing of soft sensors and soft robotic actuators. His research interests lie in digital manufacturing, sensor design, and mechanical design.
\end{IEEEbiography}

\begin{IEEEbiography}[{\includegraphics[width=1in,height=1.25in,clip,keepaspectratio]{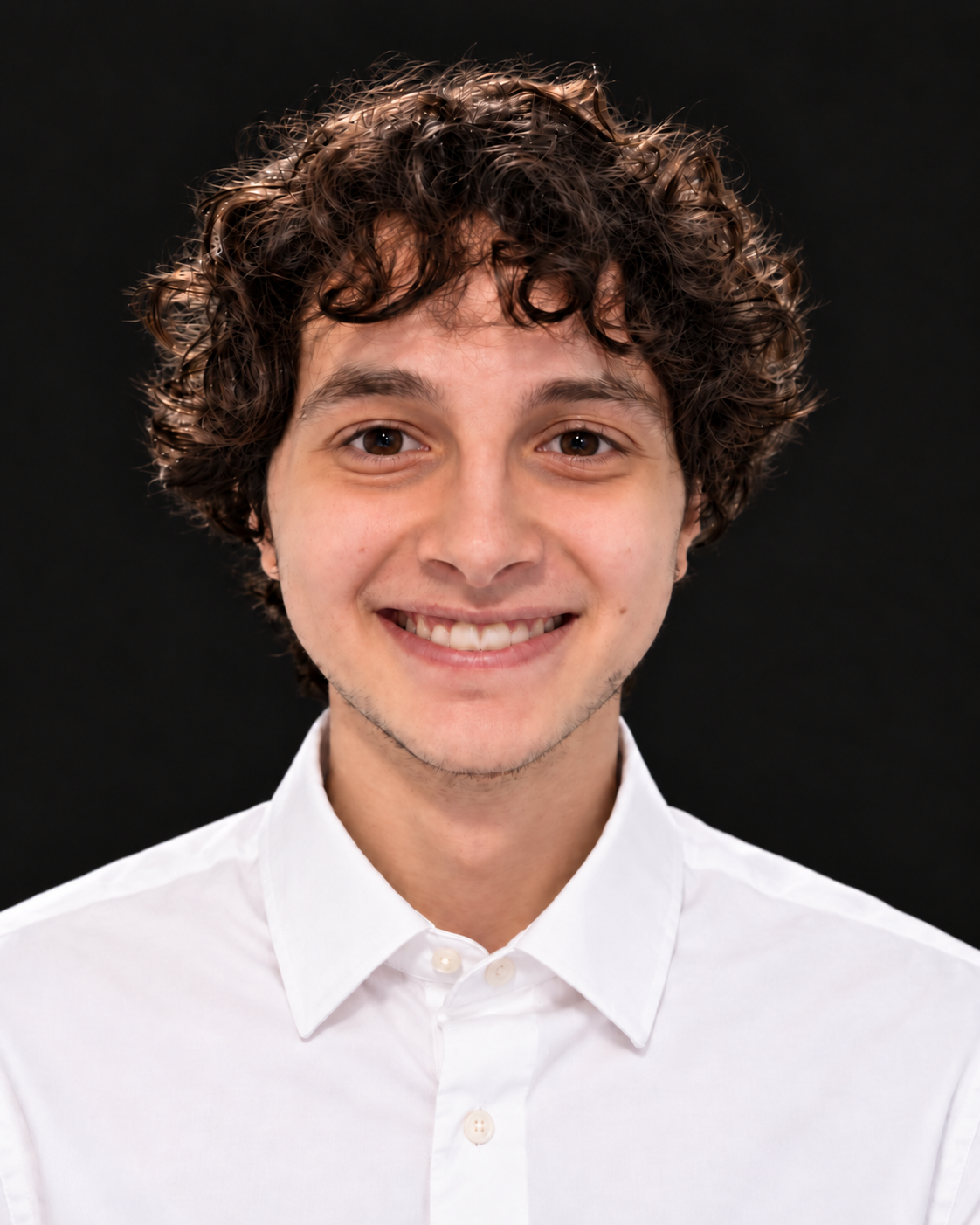}}]{Timothy Kennedy}
has been a research assistant in the Robotic Systems for Health lab for two years, supervised by Professor Jacqueline Libby. His research focuses on the development of direct-write printed sensors integrated into robotic hand rehabilitation systems. He is applying for robotics doctoral programs for Fall 2027.
\end{IEEEbiography}

\begin{IEEEbiography}[{\includegraphics[width=1in,height=1.25in, clip,keepaspectratio]{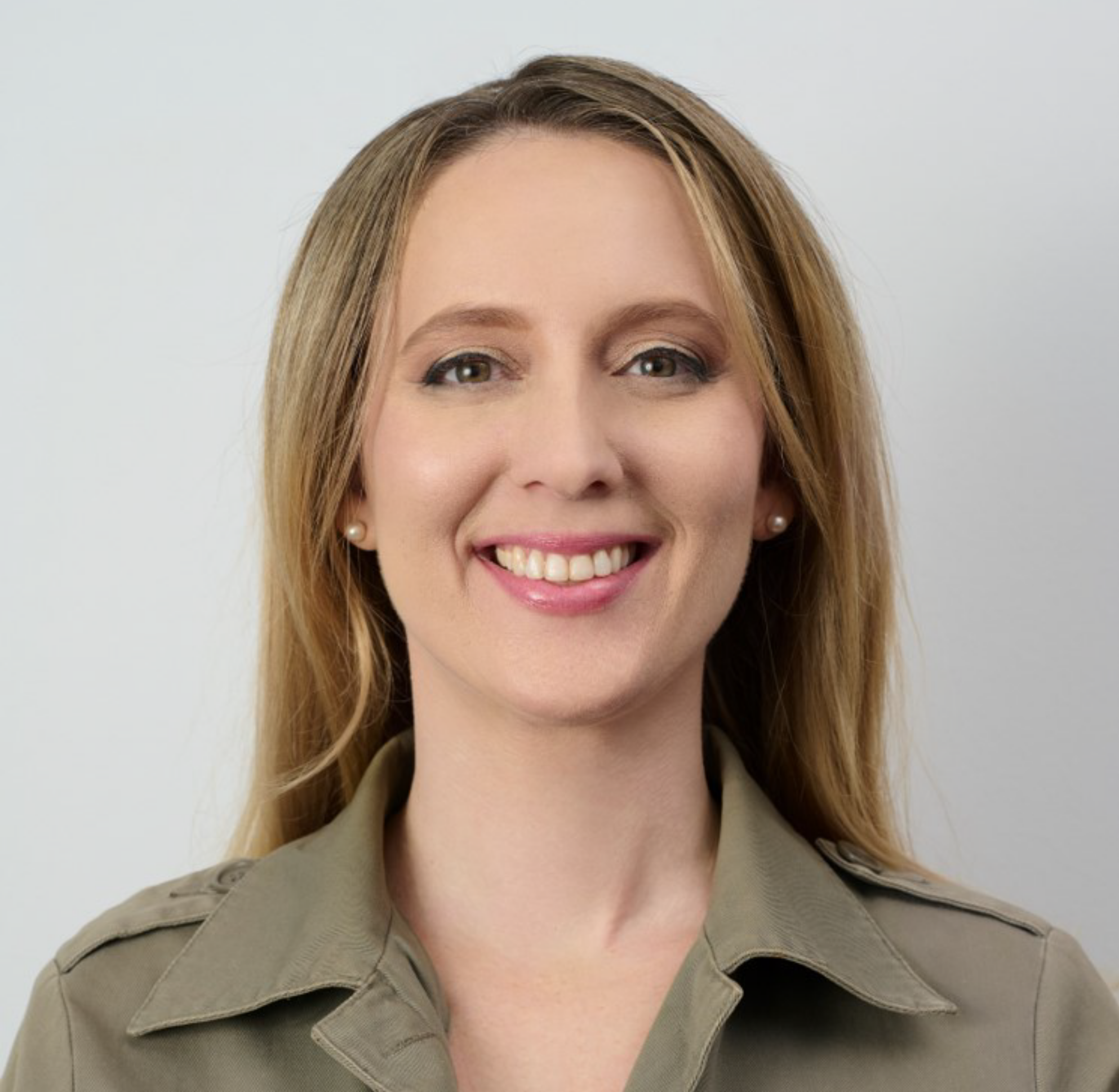}}]{Jacqueline Libby} (Member, IEEE) is an Assistant Professor at Stevens Institute of Technology in the Department of Mechanical Engineering. Her research focuses on developing soft robotic medical devices for physical therapy, rehabilitation, and pain management. She was awarded a National Science Foundation ERI grant to develop soft robotics for musculoskeletal manipulation to treat pain. Dr. Libby has an interdisciplinary background, with expertise in Robotics, Mechanical Engineering, Computer Science, and Rehabilitation Engineering. She received her B.Sc. from Brown University in Computer Science and her M.Sc. from Carnegie Mellon University in Mechanical Engineering. She earned her Ph.D. from Carnegie Mellon's Robotics Institute. Dr. Libby was awarded a two-year postdoctoral fellowship from NYU, which allowed her to conduct independent research at the intersection of engineering and rehabilitation medicine. Now at Stevens, she directs the Robotic Systems for Health Lab, where she works with a dynamic, interdisciplinary team of PhD, Master's, and Undergraduate Students.  She also teaches university courses in engineering mechanics, applied machine learning, and software engineering.
\end{IEEEbiography}

\vfill

\end{document}